\documentclass{article}

\usepackage{hyperref}
\usepackage{amssymb}
\usepackage{amsmath}
\usepackage{amsfonts}
\usepackage{graphics}
\usepackage{epsf}
\usepackage{float}
\usepackage[pdftex]{graphicx}
\usepackage{amsthm}
\usepackage{algorithm}
\usepackage{multirow}
\usepackage{authblk}
\usepackage{url}
\usepackage{psfrag}
\usepackage[lofdepth,lotdepth]{subfig}
\usepackage{xcolor}
\usepackage{booktabs}
\usepackage{cleveref}

\newtheorem{remark}{Remark}
\DeclareMathOperator*{\argmin}{arg\,min}

\begin{document}

\title{\bf Continuously heterogeneous hyper-objects in cryo-EM and 3-D movies of many temporal dimensions}


\author{Roy R. Lederman\thanks{roy@math.princeton.edu. The Program in Applied
 and Computational Mathematics, Princeton University, Princeton, NJ, USA} ~ and Amit Singer\thanks{amits@math.princeton.edu. The Department of Mathematics and the Program in Applied and Computational Mathematics, Princeton University, Princeton, NJ, USA} }

\maketitle

\begin{abstract}
Single particle cryo-electron microscopy (EM) is an increasingly popular method for
determining the 3-D structure of macromolecules from noisy 2-D  images of single macromolecules
whose orientations and positions are random and unknown.
One of the great opportunities in cryo-EM is to recover the structure of macromolecules
in heterogeneous samples, where multiple types or multiple conformations are
mixed together.
Indeed, in recent years, many tools have been introduced for the
analysis of multiple discrete classes of molecules mixed together in a
cryo-EM experiment. 
However, many interesting structures have a continuum of conformations
which do not fit discrete models nicely; the analysis of such continuously
heterogeneous models has remained a more elusive goal.
In this manuscript we propose to represent heterogeneous molecules and similar
structures as higher dimensional objects.
We generalize the basic operations used in many existing reconstruction algorithms,
making our approach generic in the sense that, in principle, existing algorithms can
be adapted to reconstruct those higher dimensional objects. 
As proof of concept, we present a prototype of a new algorithm which we use
to solve simulated reconstruction problems.

\end{abstract}


%
%
%
\section{Introduction}\label{sec:intro}

Cryo-EM has been named Method of the Year 2015 by the journal Nature Methods due to the breakthroughs that the method facilitated in mapping the structure of molecules that are difficult to crystallize. Cryo-EM does not require crystallization necessary for X-ray crystallography, and unlike NMR it is not limited to small molecules.
One of the additional great opportunities in cryo-EM, which has been noted, for example, in the surveys accompanying the Nature Methods announcement \cite{doerr2016single,nogales2016development,glaeser2016good}, is to overcome heterogeneity in the sample:
in practice many samples contain two (or more) distinct types of molecules (or different conformations of the same molecule);  methods like  X-ray crystallography and NMR, which measure ensembles of particles, have a difficulty distinguishing between these different types.
Many existing algorithms for the analysis of cryo-EM experimental data address
the problem of recovering a finite number of distinct structures in heterogeneous
samples.
However,  
''[large macromolecular assemblies] tend to be flexible, and although classification methods have come a long way 
when applied to biochemical mixtures or well-defined conformational states, the continuous motions seen for certain
samples  will challenge classification schemes and set a limit to achievable resolution, at least for a number of years.''\cite{nogales2016development}
The purpose of this manuscript is to propose an approach to mapping continuously heterogeneous structures,
and to demonstrate its applicability to cryo-EM.

The idea behind our approach is to generalize the tomography and cryo-EM problems of
recovering a function over $\mathbb{R}^3$, representing a 3-D object, 
to the problem of recovering of a function over $\mathbb{R}^3
\times \mathcal{T}$, where $\mathcal{T}$ captures the topology of the
heterogeneity. We refer to these more general functions as {\em{hyper-volumes}}, {\em hyper-objects} or {\em hyper-molecules}.
For example, loosely speaking, some types of continuous heterogeneity can be
represented  as functions over $\mathbb{R}^4$;  tomography and cryo-EM
can be generalized to the problem of recovering such 4-D objects.
Indeed, the concept of 4-D tomography (4DCT)
has been used in CT scans of patients,
whose bodies change periodically with time as they breath (see, for example, \cite{low2003method}), but it is restricted to the
case where both the orientation of each projection and its phase (time) in the breathing cycle are known.
We discuss general topologies of heterogeneity, the case of unknown
directions and ``time,'' and new approaches to the representation
and regularization of the problem.

\begin{figure}[p!]
\begin{center}
\includegraphics[width=\linewidth , trim={1in 0.5in 1in 0in},clip]{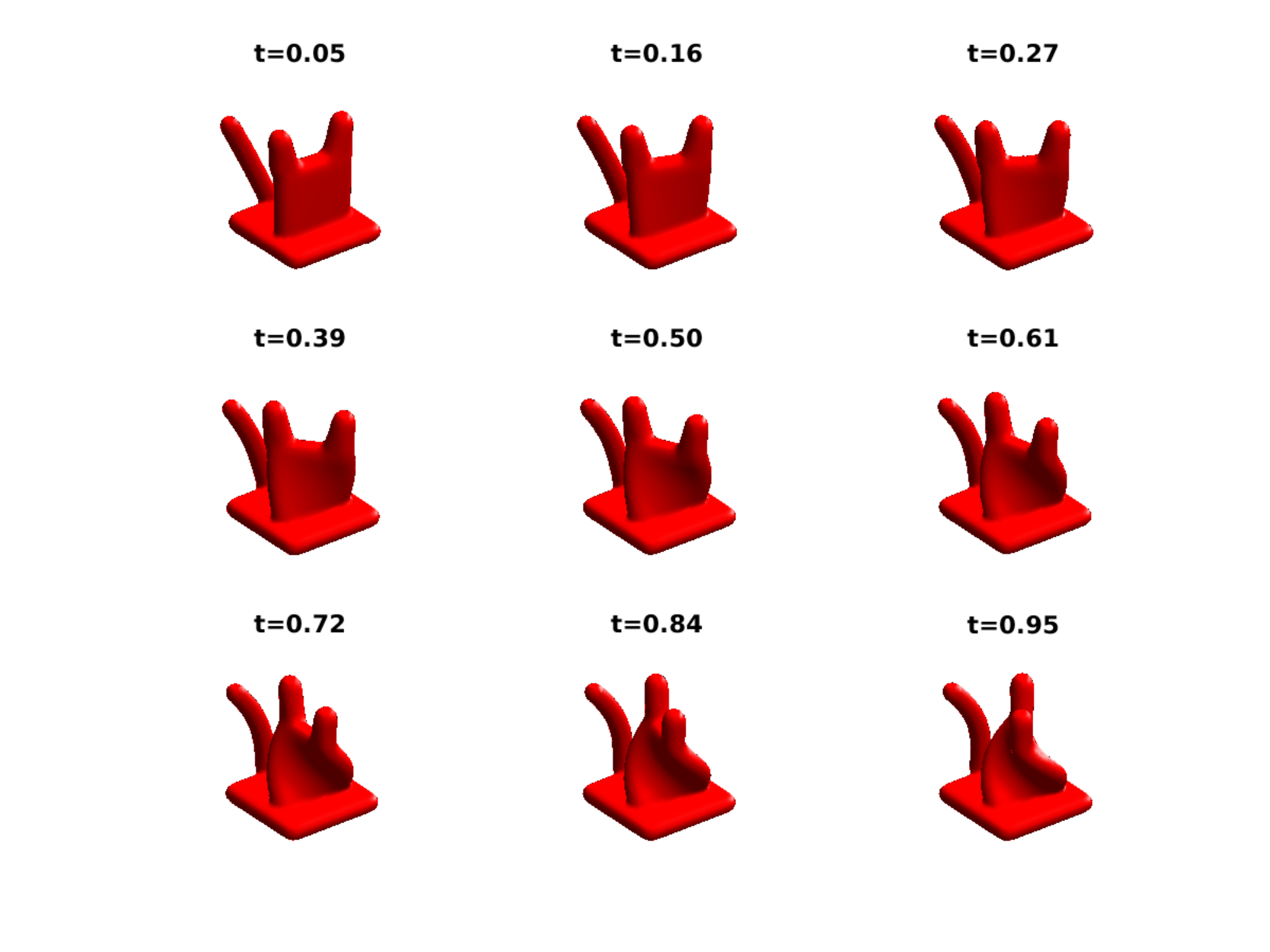}%
\end{center}
\caption{Samples of true object-instances from the simulated dataset at different values of the continuous parameter $t$, plotted at a level set } \label{fig:sim}
\end{figure}

\begin{figure}[p!]
\begin{center}
\includegraphics[width=\linewidth , trim={1in 0.5in 1in 0in},clip]{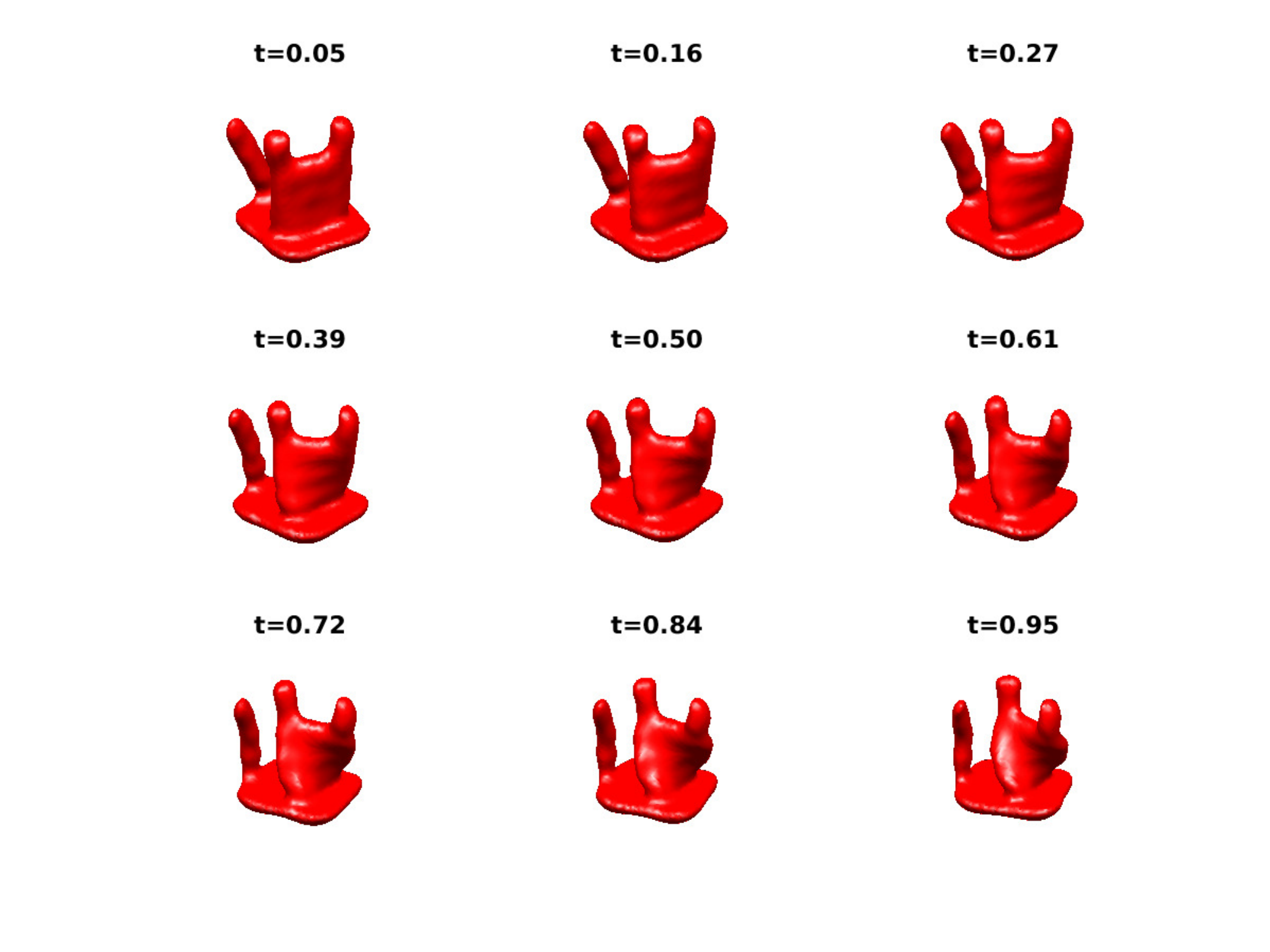}%
\end{center}
\caption{Samples of reconstructed object-instances at different values of the continuous $t$.} \label{fig:res}
\end{figure}

To illustrate the idea, we consider the heterogeneous object whose level sets are depicted in Figure \ref{fig:sim}.
This object has a one dimensional heterogeneity variable in the range $t \in [0,1]$,
so for each value of $t$ in this range, there is a 3-D object that is different from the
3-D object at a different value of $t$. However, the objects are continuous, in the sense
that the object at $t$ is very similar to the object at $t'$ if the difference $|t-t'|$ is small.
This idea is somewhat similar to a short 2-D video where the frame at each moment is generally
relatively similar to adjacent frames. In movies, this similarity allows us, for example, to compress
the video, because the video can be represented using much less information than a collection of
independent frames.

Going back to our heterogeneous object, suppose that we have a 3-D movie of this object as
we change the value of the parameter $t$ in time (examples available at  \url{http://roy.lederman.name/cryo-em}), and suppose that we focus on a single
``voxel'' or coordinate in 3-D space as time passes and measure the mass at this point at every moment.
Suppose that at $t=0$ none of the mass of the object is positioned at our point in space,
but as time passes, one of the ``ears'' passes through our point and than exits before $t=1$.
Then, we would see a curve, to which we could fit some polynomial;
we could  fit this polynomial even if we had only sample points in time, and we could use it to
interpolate the value of our voxel at times when we did not sample the voxel.
The mass at our voxel can be represented using the expansion
\begin{equation}
  \mathcal{V}(t)=\sum_q a_q P_q(t) .
  \end{equation}
Looking at all the voxels in our 3-D movie, we can fit such curves to each voxel, and obtain
a representation of our object (discretized in space); for every sample point $(x_1,x_2,x_3)$ in space,
we have the polynomials with the coefficients $a_q(x_1,x_2,x_3)$ of that particular point:
\begin{equation}
  \mathcal{V}(x_1,x_2,x_3,t) = \sum_q a_q(x_1,x_2,x_3) P_q(t).
\end{equation}
A very high degree polynomial would allow us to capture a very high frequency function,
but the expansion is typically truncated (or the high frequency are weighted down) to reflect smoothness
in the model and also due to the limited number of samples that are available,
and for practical computational reasons.
Now consider how a 3-D object is typically represented when there is no heterogeneity;
typically the representation is equivalent to some linear combination of coefficients
so that
\begin{equation}
  \mathcal{V}(x_1,x_2,x_3) = \sum_k h_k \phi_k (x_1,x_2,x_3),
\end{equation}
with $\phi_k (x_1,x_2,x_3)$ some spatial basis functions.
In cryo-EM, the objects are often represented as samples at grid-points in the Fourier domain,
so that the index $k$ encapsulates the three indices of points in the Fourier domain.
The expansion is truncated to a finite number of frequency samples, and often some
frequencies are penalized so that they would have lower amplitude; this reflects a preference to
smoother objects.
Combining our discussion of the object and its variability over ``time'' we obtain a generalization
of a 3-D object to a ``3-D movie,'' or hyper-object:
\begin{equation}\label{eq:movie_tensor_rep}
  \mathcal{V}(x_1,x_2,x_3,t) = \sum_{k,q} a_{k,q} \phi_k(x_1,x_2,x_3) P_q(t).
\end{equation}

One of the main difficulties in cryo-EM is that the 2-D images of the object are given without
information about the directing from which each image was taken, and the challenge is
to reconstruct the 3-D object from these images without knowledge of these direction; various algorithms
have been constructed for this difficult task.
In the case of heterogeneous objects, the challenge is to reconstruct hyper-objects
from images in the absence of both the direction of each image and the value $t$ that reflects the
``time in the 3-D movie'' or the version of the object which has recorded in that image.
Various algorithms have been constructed for the case of discrete heterogeneity, where there are  several distinct
objects (a collection of independent ``3-D still scenes'' as opposed to a continuous ``3-D movie'').
Continuously heterogeneous objects, such as the object in Figure \ref{fig:sim},
are often treated as if they were distinct independent objects;
this does not capture the continuous nature of the model and does not take advantage of this property to improve the reconstruction.
Very loosely speaking, this would be analogous to averaging a movie in 1 second windows,
then presenting these one-second averaged windows in random order.

We submit that 1) capturing the continuous nature of hyper-objects
contributes to the reconstruction and understanding of the objects,
2) the problem of reconstructing without knowing the direction and $t$ is analogous to
the problem of reconstructing without knowing the directions (we discuss some of the differences that do indeed arise), and 3) our general approach can be applied in a wide range of reconstruction
algorithms, based on this analogy.
Indeed, the reconstruction of distinct objects (discrete heterogeneity) can be viewed as
a special case of our approach.

We propose several directions for implementing the idea of using continuity or approximate continuity,
and discuss in more detail one such approach based on Equation (\ref{eq:movie_tensor_rep}).
We argue that different variations are useful for different models
(for example, spacial bases for local variability in a larger object).
Furthermore, we argue that the approach can be applied to rather arbitrary topologies of heterogeneity: the line segment in our example,
the cyclic pattern in 4DCT, two dimensional surfaces etc. 
We demonstrate our approach in a basic prototype as proof-of-concept, but emphasize that
these ideas can be implemented in any of the many approaches applied to problems like cryo-EM.
The results of the reconstruction using our prototype are demonstrated in Figure \ref{fig:res}.

This manuscript is organized as follows.
In Section \ref{sec:pre}, we summarize some standard results used in this manuscript.
In Section \ref{sec:setup}, we briefly review a simplified model of cryo-EM
and reformulate some of the tools used in cryo-EM algorithms in a way that we find useful
for  generalization.
In Section \ref{sec:analysis}, we discuss the generalization of tomography
and cryo-EM to heterogeneous objects.
In Section \ref{sec:alg},
we present a crude algorithm which we implemented to investigate one version
of the ideas presented in this manuscript. 
For the sake of completeness we briefly review some of the new ideas used in this implementation,
but argue that this is certainly not the only way to implement
the idea of hyper-object reconstruction.
Our preliminary results are presented in Section \ref{sec:res}.
A Brief summary and discussion of future work is presented in Section \ref{sec:future}.
Brief conclusions are presented in Section \ref{sec:conclusions}.

Additional examples and video visualization of the results
are available at \url{http://roy.lederman.name/cryo-em}.

\begin{remark}[Terminology: ``representation'']
Our use of the term ``representation'' in the context of this manuscript is very different from the
context in which we use the term in \cite{lederman2016representation}.
However, we have not found a better term that would avoid this confusion.
In this manuscript ``representation'' is a way of expressing a function or a problem,
typically an expansion of a function in some basis,
whereas in \cite{lederman2016representation} it is a technical representation theory term. 
These two works are independent and largely unrelated on a technical level; the conceptual relation between
the two is the motivation to treat heterogeneity as ``just another variable.'' 
\end{remark}

%
%
%

\section{Preliminaries} \label{sec:pre}

        \begin{table}[h!]\caption{Table of Notation}
          \begin{center}
            \begin{tabular}{r c p{10cm} }
              \toprule
              $ \mathcal{F}_Q \mathcal{V} $ & ~ & the Fourier transform of the function $\mathcal{V}$ in the variable $Q$\\
              $ \hat{\mathcal{M}} $ & ~ &  the Fourier transform of $V$ in spacial coordinates \\
              $ R {\mathbf x} $  & ~ &   the vector $\mathbf{x}$ rotated by $R$ \\
              $ R \circ  \mathcal{V} $ & ~ &  the function $\mathcal{V}$ rotated by $R$, so that $\left( R \circ \mathcal{V} \right)(\mathbf{x}) = \mathcal{V}(R^{-1}\mathbf{x})$  \\
              \bottomrule
            \end{tabular}
          \end{center}
          \label{tb:notation}
        \end{table}

        \subsection{Spherical harmonics and rotations}

        In this subsection we summarize some of the properties of spherical harmonics.

        The normalized spherical harmonic, denoted by $Y_l^m(\theta,\phi)$, with integer $l \ge 0$ and $-l \le m \le l$, is
        defined by the formula:
        \begin{equation}\label{eq:sphericalharmonics}
          Y_l^m(\theta,\varphi) =  \sqrt{\frac{2l+1}{4 \pi} \frac{(l-m)!}{(l+m)!} } P_l^m\left(\cos(\theta)\right) \exp(i m \varphi),
        \end{equation}
        where $P_l^m$ are the associated Legendre polynomials (see, for example  \cite{abramowitz1964handbook}). 
        The spherical harmonics are an orthonormal basis of $\mathcal{L}^2(S^2)$, so that
        \begin{equation}
          \int Y_l^m(\theta,\varphi) {\left(Y_{l'}^{m'}(\theta,\varphi) \right)}^* \sin(\theta) d\theta d\varphi = \delta_{ll'} \delta_{mm'} ,
          \end{equation}
        and the expansion of any function $f \in \mathcal{L}^2(S^2)$ on the sphere in this basis is
        \begin{equation}\label{eq:spherical:expand}
          f(\theta,\varphi) = \sum_{l=0}^\infty \sum_{m=-l}^l v_{lm} Y_l^m(\theta,\varphi),
          \end{equation}
        with the appropriate expansion coefficients $v_{lm}$.

        Any arbitrary rotation of a 2-D sphere can be represented by three Euler angles,
        we denote the rotation operator by $R(\alpha,\beta,\gamma)$.
        The expansion of a rotated function $R(\alpha,\beta,\gamma) \circ f $ is given by the formula
        \begin{equation}\label{eq:spherical:expandrot}
          \left(R(\alpha,\beta,\gamma) \circ f \right) (\theta,\varphi) = \sum_{l=0}^\infty \sum_{m=-l}^l \tilde{v}_{lm} Y_l^m(\theta,\varphi).
        \end{equation}
        where the coefficients in the expansion are
        \begin{equation}\label{eq:spherecoeffrot}
          \tilde{v}_{lm} = \sum_{m'=-l}^l \rho^{(l)}_{mm'}(\alpha,\beta,\gamma) v_{lm'} ,
        \end{equation}
        with $\rho^{(l)}(\alpha,\beta,\gamma)$ the $l$-th order of the appropriate form of the Wigner-D matrix
        for the rotation (see, for example \cite{coifman1968representations}).

        The restriction  $f (\varphi)$  of a function $f (\theta,\varphi)$ on a sphere to the ``equator'' is given by
        \begin{equation}
           f (\varphi) =  f (\pi/2,\varphi).
        \end{equation}
        It immediately follows that the Fourier expansion of the restricted function is
        \begin{equation}
            f(\varphi) = \sum_{l=0}^\infty \sum_{m=-l}^l v_{lm} Y_l^m(\pi/2,0) \exp(i m \varphi) = \sum_{m=-\infty}^\infty h_m \exp(i m \varphi).
        \end{equation}
        where
        \begin{equation} \label{eq:sphererestrict}
          h_m = \sum_{l=|m|}^\infty v_{lm} Y_l^m(\pi/2,0) .
        \end{equation}

        \subsection{Haar basis}

        The Haar wavelet mother function is defined by the formula
        \begin{equation}
          \varphi(t) = \begin{cases}
            1 & \text{for~} 0 \le t \le 1/2 \\
            -1 & \text{for~} 1/2 < t \le 1 \\
            0 & otherwise.
          \end{cases}  
        \end{equation}
        The Haar function $\psi_{n,k}(t)$ is defined by the formula
        \begin{equation}\label{eq:haar}
          \psi_{n,k}(t) = 2^{n/2}\varphi( 2^n t -k ),
        \end{equation}
        for all integer $n,k$.
        For the purpose of our discussion of a finite intervals,
        it is convenient to take $n \ge 0$ and $0 \le k \le 2^n-1$
        and add the constant function.
        Higher dimensional Haar basis functions are composed
        as the tensor product of one dimensional Haar functions.

        A truncated expansion of a function in the Haar basis
        is simply a piece-wise constant function.
        In other words, the space is divided into
        $2^K$ intervals, and the function is constant in each
        of these intervals.
        However, the Haar basis can also be thought of
        as a multiscale/tree decomposition of an interval (or hypercube),
        into sub-intervals. The multiscale tree structure of the functions
        is useful in analysis.

%
%
%

\section{Setup and reformulation} \label{sec:setup}

	\subsection{The representation of objects}\label{sec:setup:reps}

	An object or a ``volume'' $\mathcal{V}$ is a function over $\mathcal{X}$ (in our case, 
	$\mathcal{X} = \mathbb{R}^3$ is coordinates in space). 
	The Fourier transform of the object, which we denote by $\mathcal{F}_X
	\mathcal{V} = \hat{\mathcal{V}}$, is a function over $\Omega$ (in our case
	$\Omega = \mathbb{R}^3$ are coordinates in the Fourier domain).
	For the purpose of this discussion, we assume that the objects are
	continuous with respect to $X$ and $\Omega$ (this assumption can be relaxed).

	In applications, the function $\mathcal{V}$ must be discretized in some way, if only
	so that we can represent it in a digital computer. 
	Furthermore, the choice of discretization reflects (implicitly or explicitly)
	priors or regularization of the object, which are required in order to make the
	tomography problem tractable.
	
	The object $\mathcal{V}$ is often represented using discrete samples of
	$\mathbb{R}^3$ on a regular grid, restricted to some finite box which is
	sufficiently large to contain the molecule which we wish to reconstruct (e.g., 
	a $128 \times 128 \times 128$ grid); points off the grid are sometimes
	evaluated using some continuous interpolation from the grid points (although
	non-continuous voxels are also considered).
	Since many of operations in cryo-EM are naturally represented in the
	Fourier domain, and as it is often assumed that $\mathcal{V}$ (or some
	low-resolution version of it) is band-limited, many cryo-EM algorithms
        and programs
	use a similar regular grid in the Fourier domain to represent samples of
	$\hat{\mathcal{V}}$ in $\Omega$.
	In works such as \cite{barnett2016rapid}, the Fourier transform $\hat{\mathcal{V}}$ of
	the object is represented as concentric shells sampled at various radii. 
	
	These different representations assume certain properties of $\mathcal{V}$;
	the regular grid in a box in the real space assumes that the $\mathcal{V}$ is
	compactly supported in real space (the choice of interpolation scheme
	implies additional smoothness assumptions). 
	The regular grid in the Fourier domain implies, for example, that the function is
	band-limited.
	The representation in concentric shells in the Fourier domain makes similar
	assumptions on the band-limit of the function and is computationally
	convenient; each concentric shell can be represented using spherical harmonics
	rather than samples on the sphere, yielding a natural continuous representation
	that has no directional bias (unlike the regular grid).

	Loosely speaking, different common representations of an object can be
	summarized as vectors of coefficients. For example, a vector composed of all
	the values sampled at the grid points of a 3-D Fourier transform of an object.
	In standard linear representations of objects, 
        the object (or its Fourier transform) is written as
	\begin{equation}\label{eq:rep:expansion}
		\mathcal{V}(x_1,x_2,x_3) = \sum_i w_i
		\varphi_i(x_1,x_2,x_3),
	\end{equation}
        or
	\begin{equation}\label{eq:rep:fourier:expansion}
		\hat{\mathcal{V}}(\omega_1,\omega_2,\omega_3) = \sum_i w_i
		\hat{\varphi}_i(\omega_1,\omega_2,\omega_3),
	\end{equation}
        where  $\mathbf{w}$ is the vector of coefficients, 
        $w_i$ the $i$-th element of the vector,
        $\varphi_i$ some basis functions appropriate for that representation,
        and $\hat{\varphi}_i$ their Fourier transforms.

	Typically, the representation of objects is continuous, in the
	sense that small rotations and translations of the object, and small changes in
	the object, result in small changes to the coefficients in the representation.
	This property is not necessarily uniform: for example, some types of perturbation in the objects
	can have a larger effect on high frequency components than on low
	frequency components of some representations.

	The choice of basis and the truncation of the basis to a finite number of
	basis functions (e.g., finite cube in space and finite resolution) restricts the
	functions that can be represented to certain function spaces. This serves as an
	implicit regularization in algorithms for the reconstruction of molecules (see below), and it is
	sometimes augmented by more explicit regularization or priors on the object.

	Bases that are better suited to a problem allow high accuracy approximation of
	the relevant functions using fewer coefficients, or more efficient
	computation. 
	Loosely speaking, a function that is  efficiently approximated in one
	basis can be approximated in other reasonable bases even if the representation
	is less efficient and requires more basis functions. 
	For example, a low order polynomial can be represented efficiently as a linear
	combination of polynomials basis functions, and it can also be approximated
	well by equally spaced samples with linear interpolation between the sample
	points. In this sense, all the bases are ``equivalent.''
	However, an efficient basis is also an implicit regularizer which restricts the
	space of functions. For example, low order polynomials are smooth and have
	restricted oscillations, whereas the sampling scheme above has more degrees of
	freedom and represents functions that are not smooth and functions that
	oscillate more.
	Such restriction can be introduced in any arbitrary representation explicitly
	as constraints, filters, penalties or regularizers.

	\subsection{Cryo-EM}

        Electron Microscopy is an important tool for recovering the 3-D structure of molecules.
        Of particular interest in the context of this manuscript is Single Particle Reconstruction (SPR), 
        and more specifically, cryo-EM,
        where multiple noisy 2-D projections, ideally of identical particles in different orientations,
        are used in order to recover the 3-D structure.
        The following formula is a simplified noiseless imaging model of SPR, for obtaining
        the noiseless image $I^{(i)}$ from a object $\mathcal{V}$ (representing the molecule's density):
	\begin{equation} \label{eq:cryoemmodel:space}
		I^{(i)}(x_1,x_2) = \left( \mathcal{P} R_i \mathcal{V} \right)(x_1,x_2) =
		\int_{\mathbb{R}} \mathcal{V}(R_i^{-1} {\mathbf x} ) d x_3 ,
	\end{equation}
	where ${\mathbf x}=(x_1,x_2,x_3)^\intercal$ and $R_i$ is the rotation that determines the orientation of the molecule.
        In other words, the model is that the molecule is rotated in a random direction, and the recorded image is the top-view projection of the rotated molecule, integrating out the $x_3$ or $z$ axis.
        In the Fourier domain, the Fourier transform $\hat{I}^{(i)}$ of an image is a
	slice  of the Fourier transform $ \hat{\mathcal{V}}$ through the origin:  
	\begin{equation} \label{eq:cryoemmodel:freq}
		\hat{I}^{(i)}(\omega_1,\omega_2) = \hat{\mathcal{V}} (R_i^{-1} {\mathbf \omega} )
	\end{equation}
	where ${\mathbf \omega} = (\omega_1,\omega_2,0)^\intercal$ and $R_i$ is the random rotation.

        One of the characteristic properties of cryo-EM SPR that sets it apart from other tomography techniques is that the orientation $R_i$ of the molecule in each image is unknown in cryo-EM, whereas in other tomography techniques the rotation angles are typically recorded with the measurements.

\begin{figure}
          \begin{center}
            \includegraphics[width=1.5in,trim={0.3in 0.15in 0 0},clip]{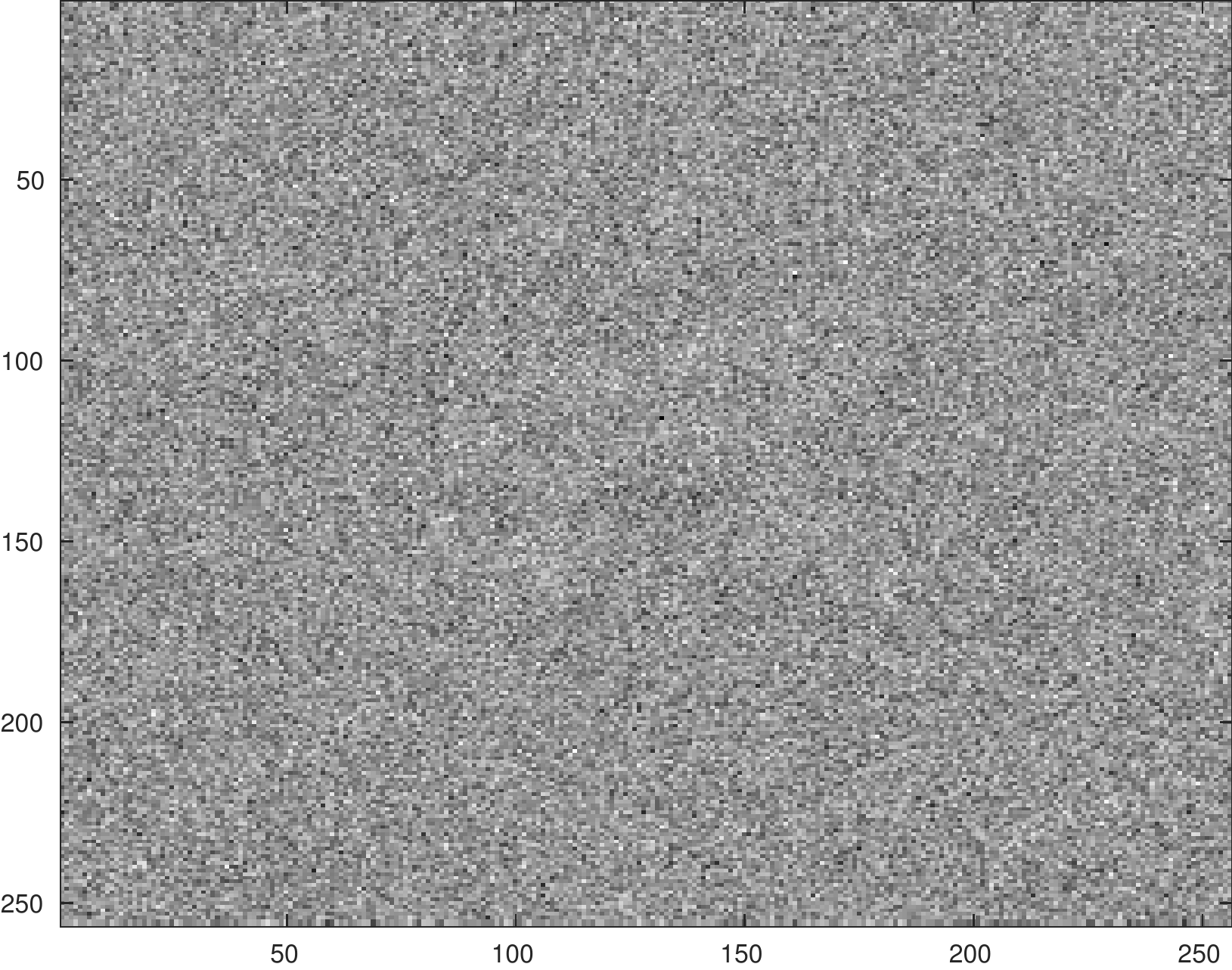}%
            ~\includegraphics[width=1.5in,trim={0.3in 0.15in 0 0},clip]{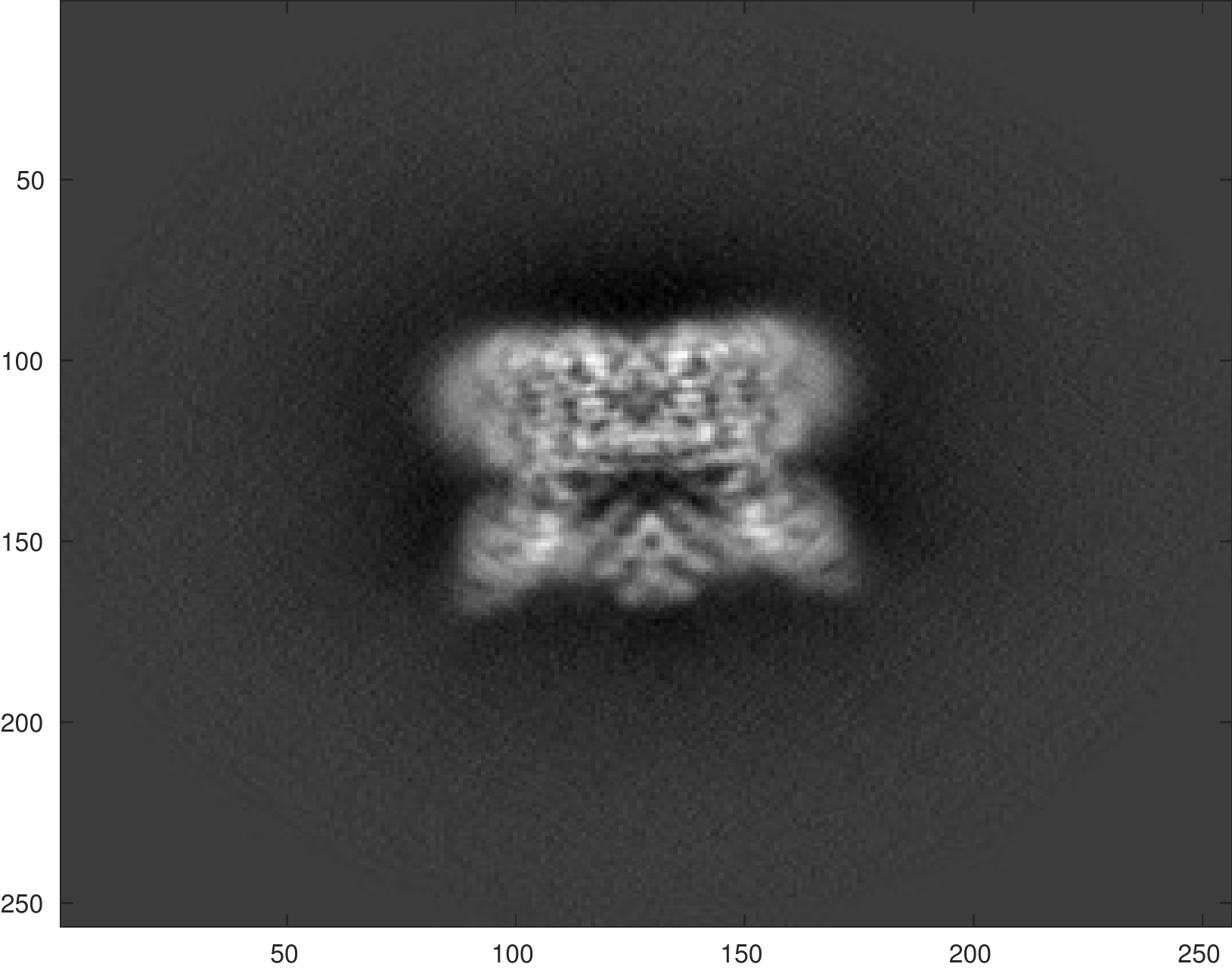}%

            \includegraphics[width=1.5in,trim={0.3in 0.15in 0 0},clip]{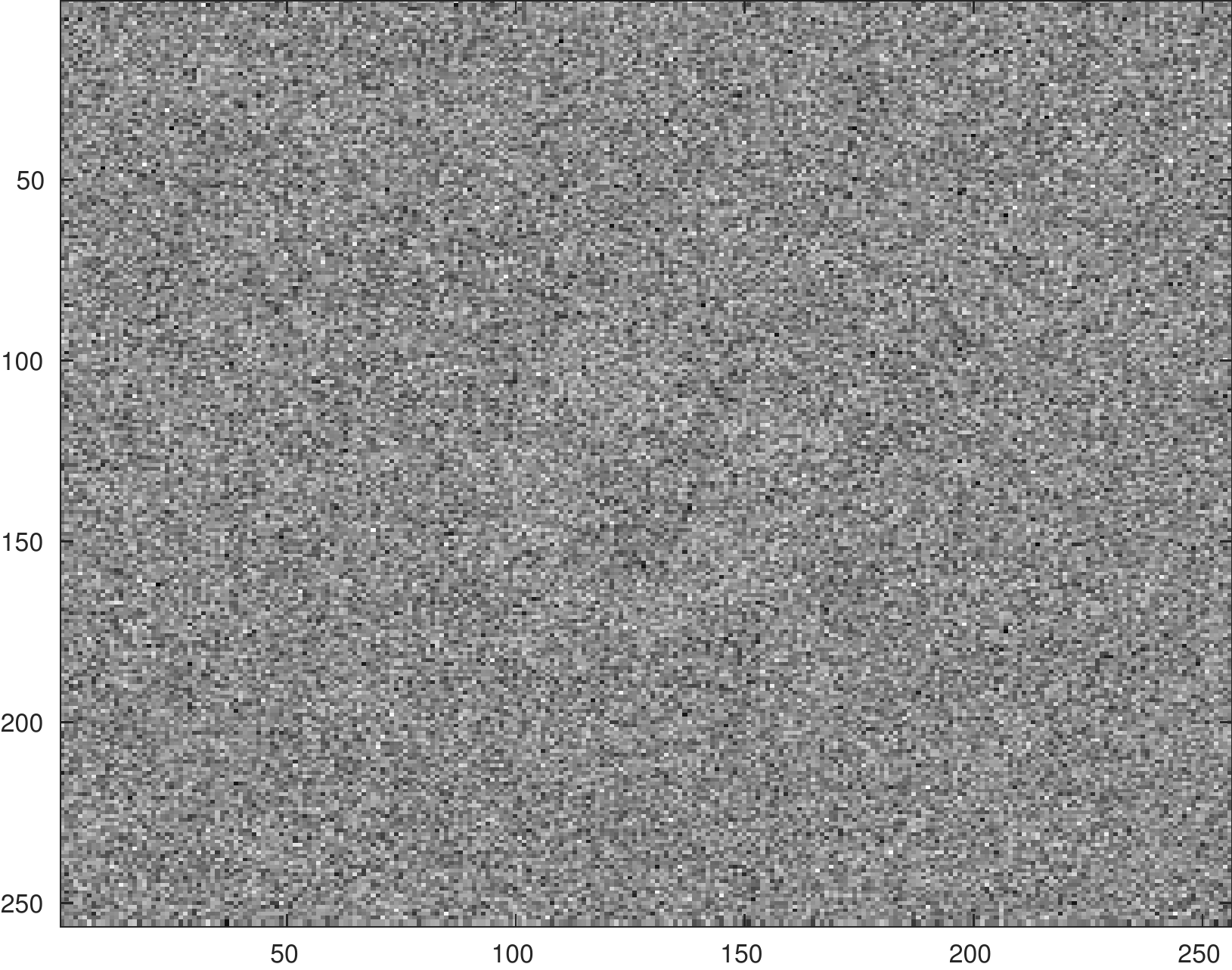}%
            ~\includegraphics[width=1.5in,trim={0.3in 0.15in 0 0},clip]{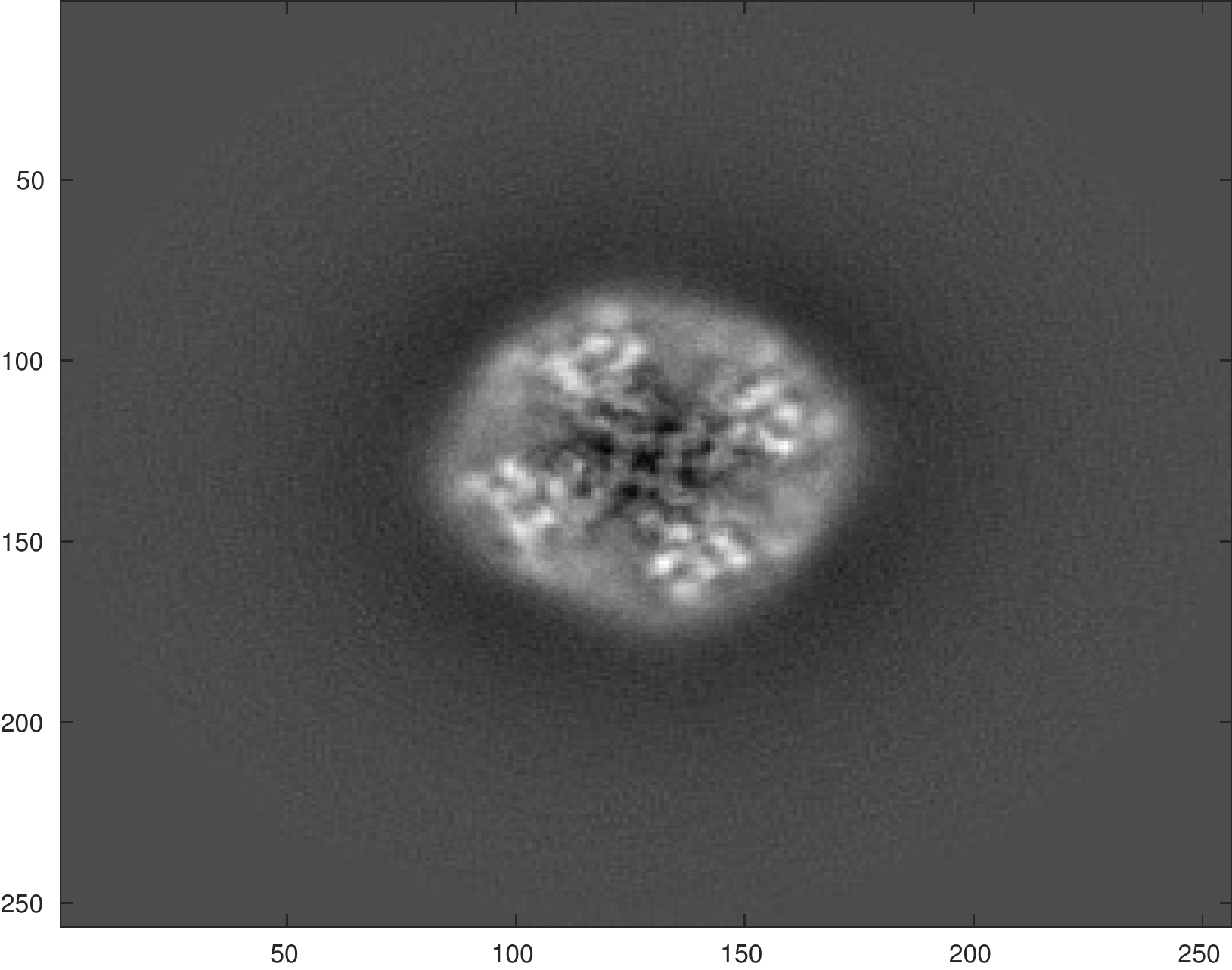}%
          \end{center}
          \caption{Left: two raw experimental images of  TRPV1, available via EMDB 5778 \cite{liao2013structure}. Right: computed projections of TRPV1 which are the closest to the images on their left. }\label{fig:cryosample}
        \end{figure}

        The analysis of cryo-EM images is further complicated by many additional effects,
        some of the most notable are:
	\begin{itemize}
	  \item extremely high levels of noise, far exceeding the signal in magnitude (see sample images in Figure \ref{fig:cryosample}),
	  \item filters associated with the imaging process (CTF),
	  \item an unknown shift of each image in the image plane,
	  \item and the discretization of the measurements.
	\end{itemize}
        More detailed discussions of these challenges, and various other challenges
        can be found, for example, in \cite{frank2006three}.
	Since these aspects are studied extensively in other works, and since the goal
	of this work is to introduce a new generalization of previous works,
        we will not discuss these aspects in much detail when they do 
	not raise too many new issues that are of particular interest in continuous heterogeneity
        at the level of this preliminary discussion.
	We note that in our experiments we use simulated data with high levels of noise.

	\subsection{The tomography problem in cryo-EM}
	
	Suppose that we are given a set of images $\{I^{(i)}\}_{i=1}^n$ with the
	orientation $\{R_i\}_{i=1}^n$ of each image. 
	Then, the reconstruction problem in cryo-EM is a classic tomography problem.
	In the Fourier domain, the model in equation (\ref{eq:cryoemmodel:freq})
	suggests that the problem is to reconstruct $\hat{\mathcal{V}} $ from
	(very noisy) slices of the Fourier transform of the object. 
	This problem is ill-posed even in the noiseless case since we have no
	information about the values of $\hat{\mathcal{V}}$ at points between the
	slices. However, additional priors and assumptions on the properties of the object
	make the problem more tractable. 
	
	For the purpose of our discussion of tomography, we propose a further
	simplified description of the tomography problem, a {\em one frequency pixel model} of tomography,
	where we receive scalar samples $S_i$ of $ \hat{\mathcal{V}}$, each at a
	given frequency point $ \omega^i =
	(\omega^i_1,\omega^i_2,\omega^i_3)^\intercal$:
	\begin{equation} \label{eq:onepixletomography}
	  S^{i} = \hat{ \mathcal{V}}(\omega^i_1,\omega^i_2,\omega^i_3).
	\end{equation}
	It is easy to see that if the Fourier transform of each image $\hat{I}^{(i)}$ in
	(\ref{eq:cryoemmodel:freq}) is considered to be a (finite) collection of samples
	of $ \hat{\mathcal{V}}$ at various frequency points, then the
	one frequency pixel model (\ref{eq:onepixletomography}) describes the
	same model as (\ref{eq:cryoemmodel:freq}), with the latter grouping together
	samples into images.

	Taking into account the discussion of representing an object as a vector $\mathbf{w}$
	of coefficients of some arbitrary linear expansion (\ref{eq:rep:expansion}),
	we propose a more general {\em linear operator formulation} of linear
	models such as (\ref{eq:cryoemmodel:space}),
	(\ref{eq:cryoemmodel:freq}) or (\ref{eq:onepixletomography}), which simply
	states that each (noiseless) measurement is some vector of coefficients $\mathbf{y^{(i)}}$ given by a
	linear operation of the vector of coefficients $\mathbf{w}$:
	\begin{equation} \label{eq:operatortomography}
	  \mathbf{y^{(i)}} = A^{(i)} \mathbf{w} .
	\end{equation}
	In the cryo-EM imaging model, the linear operator $A^{(i)}$ captures
	the operation of rotating a molecule, projecting it and representing the
	resulting noiseless image using the vector of coefficients $\mathbf{y^{(i)}}$
        (more comprehensive practical models also take into account
        the CTF, shifts, discretization, noise, etc.).

	\subsection{Unknown orientations, and reconstruction algorithms in
		cryo-EM SPR}
	
	One of the characteristics of the cryo-EM SPR problem that sets
	it apart from other tomography problems, is that the orientations $R_i$ of each
	image is unknown, making the problem severely ill-posed.

	For the purpose of this discussion, we make a simplistic distinction
	between two approaches to the cryo-EM reconstruction problem:
	object-free image alignment, and direct object estimation.
        The object-free image alignment approach typically 
        takes advantage of the projection-slice theorem to estimate the rotation
        $R_i$ of each image with respect to others based on the intersection
        of slices in the Fourier domain
        (see, for example, \cite{van1987angular,singer2010detecting,singer2011three,shkolnisky2012viewing}).
        Once the rotations are recovered, the next step is the tomography problem of 
	estimating the object given the rotations.
	In the current manuscript, we focus on methods of direct object estimation, 
	often called {\em{refinement}},
	which alternate between refining an estimate of the object, and
	estimating the rotation of each image with respect to the object:
        given an estimated object $\mathcal{V}^{(n)}$ at step $n$, estimate the most likely
        orientation $R_{n,i}$ of the image recorded in the $i$-th image, for all images:
        \begin{equation}\label{eq:refine:orientation}
          R_{(n,i)} = \argmin_{R} \|A(R) \mathcal{V}^{(n)} - I^{(i)}\| ,
        \end{equation}
        where $A(R)$ is the operator which produces an image from the object
        $\mathcal{V}^{(n)}$, rotated to orientation $R$, and $I^{(i)}$ is the
        $i$-th recorded image.
        In practice, this minimization is typically performed by comparing
        each image to a set of computed template projections of the estimated object at some sampled values of $R$. 
        Given these estimated rotations, produce a new estimate of the object, for example by solving:
        \begin{equation}\label{eq:refine:volume}
          \mathcal{V}^{(n+1)} = \argmin_{\mathcal{V}} \sum_{i=1}^N \|A(R_{(n,i)}) \mathcal{V} - I^{(i)}\|^2 ,
        \end{equation}
        possibly with some additional regularization terms.
        Some popular refinement software packages and algorithms follow more elaborate statistical approaches
        (e.g., \cite{scheres2005maximum,sigworth2010chapter,grigorieff2007frealign,scheres2012bayesian,scheres2012relion,punjani2016building}) such as the Bayesian approach, and other optimization schemes (e.g., stochastic gradient descent in \cite{punjani2017cryosparc}).
        Since these approaches use similar fundamental operators and due to the limited scope of this
        manuscript, we will restrict our attention to conventional refinement algorithms and argue
        that our approach generalizes such algorithms to hyper-molecules.
        Our approach can be used in a similar way to generalize more elaborate algorithms.

	\subsection{Frequency marching}

        Several refinement tools such as RELION\cite{scheres2012bayesian,scheres2012relion}
        and FREALIGN \cite{grigorieff2007frealign, grigorieff2016chapter, lyumkis2013likelihood}
        gradually increase the resolution of the estimated object as they iterate.
        In \cite{barnett2016rapid}, this concept is reformulated as {\em frequency marching}:
        starting the iterative process with a representation of the object that uses
        only a small number of low frequency basis functions, and adding higher
        frequency basis functions to the expansion at subsequent refinement iterations.

        Frequency marching highlights the fact that use of basis functions
        is not merely an technical implementation detail of how to store
        a function digitally, but rather a key tool at the heart of the algorithm.

	\subsection{Heterogeneity}

        So far, we have assumed that all the molecules being imaged in an experiment are identical copies of each other, so that all the images are projections of identical copies of the object $\mathcal{V}$, from different directions. 
        However, in practice, the molecules in a given sample may differ from one another for various reasons. 
        For example, the sample may contain several types of different molecules due to some contamination or feature of the experiment. Alternatively, the molecules which are studied may have several different conformations or states, or some local variability. 
        The heterogeneity may be discrete (e.g., in the case of distinct different molecules), continuous (in the case of molecules with continuous variability), or a mixture of continuous elements and discrete elements.

        In the heterogeneous settings, 
        the simplified noiseless imaging model of the
        homogeneous case (see (\ref{eq:cryoemmodel:space}))
        is generalized to 
      	\begin{equation} \label{eq:cryoemmodel:het:space}
	  I^{(i)}(x_1,x_2) = \left( \mathcal{P} R_i \mathcal{V}[t_i] \right)(x_1,x_2) =
	  \int_{\mathbb{R}} \mathcal{V}[t_i](R_i^{-1} {\mathbf x} ) d x_3 ,
	\end{equation}
        where $\mathcal{V}[t_i]$ is the objects imaged in sample $i$. 
        The generalized operator formulation (\ref{eq:operatortomography}) is
       	\begin{equation} \label{eq:het:operatortomography}
	  y^{(i)} = A^{(i)} \mathbf{w}^{(t_i)} ,
	\end{equation}
        where $\mathbf{w}^{(t_i)}$ is the vector of coefficients in the representation
        of the object $\mathcal{V}[t_i]$.

        Most existing algorithms and software tools treat only the case of discrete heterogeneity.
        In conventional refinement algorithms,
        discrete classes of molecules are
        reconstructed by modeling multiple independent objects (see, for example \cite{van1985characteristic});
        given multiple estimated objects $\{ \mathcal{V}^{(n)}[l] \}_{l=1}^L$,
        we estimate the most likely pair of class $l_{(n,i)}$ and orientation $R_{(n,i)}$
        for each images, and proceed to produce a new estimate of each objects based on the
        the images assigned to it.
        For example, the generalization of (\ref{eq:refine:orientation}) and (\ref{eq:refine:volume}) is
        \begin{equation}\label{eq:refine:het:orientation}
          (R_{(n,i)},l_{(n,i)}) = \argmin_{(R,l)} \|A(R) \mathcal{V}^{(n)}[l] - I^{(i)}\| ,
        \end{equation}
        and
        \begin{equation}\label{eq:refine:het:volume}
          \mathcal{V}^{(n+1)}[l] = \argmin_{\mathcal{V}} \sum_{i: l_i = l} \|A(R_{(n,i)}) \mathcal{V} - I^{(i)}\|^2 ,
        \end{equation}        
        A similar generalization is used in more elaborate algorithms
        (e.g., \cite{scheres2005maximum,scheres2012bayesian})
        and in software packages such as RELION\cite{scheres2012relion}.
        Other approaches to heterogeneity require some method of recovering the rotation of the images
        although the images reflect mixtures of projections of different molecules
        (e.g., \cite{katsevich2015covariance,anden2015covariance}).
        We treat heterogeneity in object-free algorithms in \cite{lederman2016representation} (with generalization 
	to continuous heterogeneity - in preparation). 
        Another recent independent work on object-free reconstruction \cite{aizenbud2016max} proposes to iterate between
        estimating the orientations and estimating the class labels based on pairwise relations between images. 
        A different perspective on object-free algorithms for continuous heterogeneity,
        proposed in \cite{schwander2014conformations,frank2016continuous},
        is to construct a manifold of images and study the low dimensional structures
        induced by rotations and heterogeneity.

        Currently, the prevailing approach to continuous heterogeneity is to treat the object
        roughly as if it were a collection of discrete independent objects, rather than
        a continuum of objects.  In this manuscript manuscript we investigate whether we can capture
        the continuous nature of the states, and even take advantage of it
        in reconstruction.

%
%
%
\section{Analytical apparatus}\label{sec:analysis}

	\subsection{Heterogeneous molecules - ``hyper-molecules''}\label{sec:analysis:het:hyper}

	The purpose of this section is to generalize the definition of an object to
	a hyper-object, which represents a heterogeneous set of objects, simply by adding a
	variable that identifies each object instance.

	To generalize the definition of an object to an hyper-object,
	we define a family of objects $\mathcal{M}$ as a function over $\mathcal{X} \times
	\mathcal{T}$, where $\mathcal{T}$ is the index set or parameterization of the
	family of objects.
	The  evaluation of the density of an object at the
	coordinates $(x_1,x_2,x_3)$, is analogous to the evaluation of
        a hyper-object at the coordinates
	$((x_1,x_2,x_3),t)$, which means choosing the object instance with the index $t$ and
	evaluating the density of that object at the coordinates $(x_1,x_2,x_3)$.	
	To illustrate the notation in the discrete case, suppose that we have only two distinct
	objects, then $\mathcal{T} = \{ 0,1 \}$,  
	and the object instances evaluated at $(x_1,x_2,x_3) \in \mathcal{X}$ are 
	$\mathcal{M}((x_1,x_2,x_3),0)$ and $\mathcal{M}((x_1,x_2,x_3),1)$.
	In the continuous case, suppose that we parameterize the hyper-objects by 
	$\mathcal{T} = [0,1]$, then 
	the object at $t \in \mathcal{T}$ evaluated at $(x_1,x_2,x_3)$ is
	$\mathcal{M}((x_1,x_2,x_3),t)$.
	We note that, in general,  $\mathcal{T}$ can have various topologies (discrete,
	an interval, a subset of a multi dimensional space, a torus, combinations
	thereof, etc.).
        For example, in 4DCT, the topology of heterogeneity may capture the cyclic behavior
        of breathing (as opposed to an interval with independent ends);
        in cryo-EM it may capture a one dimensional variability in states,
        or, for example, independent local variability in two different areas of the molecule.

	We denote by $\mathcal{F}_X$ the Fourier transform which integrate over
	the coordinate space $\mathcal{X}$, and does not interact with the
	parameterization $\mathcal{T}$, so the Fourier transform of the object indexed by
	$t\in\mathcal{T}$, and evaluate at the frequency
	$(\omega_1,\omega_2,\omega_3)$ is $\hat{\mathcal{M}}
	((\omega_1,\omega_2,\omega_3) ,t) = \left(\mathcal{F}_X \mathcal{M}\right)
	((\omega_1,\omega_2,\omega_3) ,t)$.

	Finally, we define a transform $\mathcal{L}_{T}$ on the
	parameterization $\mathcal{T}$ which is analogous to $\mathcal{F}_X$.
	For example, suppose that $\mathcal{T} = \{0, 1, \ldots , M-1\}$, 
	and suppose that  $\mathcal{L}_{T}$ is the discrete Fourier transform.
	Then, we have the ``parameter-wise'' discrete Fourier transform of the object
 	evaluated at {\em parameter frequency} $\tau$ and coordinates $(x_1,x_2,x_3)$,
	$\left( \mathcal{L}_{T} \mathcal{M} \right)((x_1,x_2,x_3),\tau)$.
	In this case,  $\tau=0$ yields the average of all objects:  
	$\left( \mathcal{L}_{T} \mathcal{M} \right)((x_1,x_2,x_3), 0) \propto
	\frac{1}{M} \sum_{t=0}^{M-1} \mathcal{M}((x_1,x_2,x_3),t)$.
	We may also apply both transforms: $\left( \mathcal{L}_{T}
	\mathcal{F}_X \mathcal{M} \right) ((\omega_1,\omega_2,\omega_3) ,\tau)$.
	For the purpose of this discussion, we assume that the 
        operators commute: \\ $
		\left( \mathcal{L}_{T} \mathcal{F}_X \mathcal{M}
		\right) ((\omega_1,\omega_2,\omega_3) ,\tau) = \left(
		\mathcal{F}_X \mathcal{L}_{T} \mathcal{M} \right)
		((\omega_1,\omega_2,\omega_3) ,\tau)$.

	\subsection{Tomography of hyper-objects}

	The caricature one frequency pixel model of tomography
	(\ref{eq:onepixletomography}) describes tomography as the inverse problem of
	recovering the Fourier transform of a object $\mathcal{V}$ from many noisy
	samples $\{I^i\}_{i=1}^n$ at known frequency coordinates
	$\{(\omega^i_1,\omega^i_2,\omega^i_3)\}_{i=1}^n$ (with some constraints or
	regularization to make the inverse problem tractable).
	
	This description of the tomography problem generalizes naturally to
	hyper-objects: the one frequency pixel model of hyper-object tomography is
	the inverse problem of recovering $\mathcal{F}_X \mathcal{M}$ from many
	noisy samples $\{I^i\}_{i=1}^n$ at known frequencies and values of the
	heterogeneity index $\{((\omega^i_1,\omega^i_2,\omega^i_3),t_i)\}_{i=1}^n$,
	with some analogous constraints or regularization of the hyper-object. 
	In the discrete case, this simply means that for each sample, we have the
	label  $t_i$ of the the object which was measured, and the frequency
	coordinates of the sample $(\omega^i_1,\omega^i_2,\omega^i_3)$, and we may
	collect all the samples of each object and proceed to process each object
	independently from the others.

        In computed tomography (CT), 4-D tomography technology\cite{low2003method} (4DCT) has
        been proposed in order to image the body of a patient in different
        states of the breathing cycle, so that
        $\mathcal{T}$ is time or phase within a breathing cycle.
        In 4DCT, as in classic CT, the orientation of each image is known;
        in addition, the position of each image in the breathing cycle
        is recorded using external means.
        The original 4DCT reconstruction algorithms binned the images according to
        their position in the cycle and then reconstructed an independent
        objects from each discrete bin.
        In recent years, regularization techniques have been introduced in order to
        take advantage of the relation between the volumes in different
        phases of the cycle
        (see, for example, \cite{jia20104d,gao20124d,kazantsev2015employing,gopi2014multiple}).

	\subsection{The representation of heterogeneous objects - the discrete case}

	When there are K classes of objects, they can be represented as K 
	vectors of coefficients. 	 
	Suppose that $\mathbf{w}^{(k)}$ is the vector of coefficients associated with the
	representation of the $k$-th object, with ${w}^{(k)}_i$ the $i$-th element of
	this vector. 
	Obviously, we can rewrite these as columns of a matrix $W$, such that
	$W_{ik}={w}^{(k)}_i$, and the $k$-th object as
	\begin{equation}\label{eq:rep:het:disc}
		\hat{\mathcal{M}}((\omega_1,\omega_2,\omega_3),k) = \sum_i W_{ik}
		\varphi_i(\omega_1,\omega_2,\omega_3) .
	\end{equation}
        
        Clearly, ${\mathcal{M}}((x_1,x_2,x_3),k)$ is equivalent
        to $\mathcal{V}[k](x_1,x_2,x_3)$ used in (\ref{eq:cryoemmodel:het:space}),
        so that in the discrete case this formulation of
        the problem is equivalent to the classic formulation
        used in existing cryo-EM algorithms which store multiple
        independent models of objects.

	As a step toward the discussion below of representations of continuously
	heterogeneous objects, we present some non-formal motivation through the
	following trivial accounting exercise.
        Suppose that $U$ is a $K \times K$ unitary matrix, such as the Discrete
	Fourier Transform matrix (this requirement can be relaxed in various ways with
	minor modifications to the discussion).
	We define the matrix $\tilde{W} = W U$, where each column is a linear
	combination of columns in $W$.
	The $k$-th objects is now written as
	\begin{equation}
		\hat{\mathcal{M}}((\omega_1,\omega_2,\omega_3),k) = \sum_i \sum_l \tilde{W}_{il}
		U^{-1}_{lk} \phi_i(\omega_1,\omega_2,\omega_3) .
	\end{equation}	
	Clearly, we can choose the first column of $U$ to be constant, obtaining the
	the average of all molecules 
	\begin{equation}
		\overline{\hat{\mathcal{M}}}(\omega_1,\omega_2,\omega_3) = \sum_i \tilde{W}_{i1}
		\varphi_i(\omega_1,\omega_2,\omega_3)  = \frac{1}{n} \sum_k
		\hat{\mathcal{M}}((\omega_1,\omega_2,\omega_3),k) .
	\end{equation}	
	Generally speaking, we do not have a preference for the orientation of an object
	that we represent (with some technical exceptions). Similarly, when we
	represent several classes of objects, we do not have any preference
	for the relative orientation of one class with respect to the others. 
	However, the trivial example above demonstrates that when we have
	multiple classes of objects that are somewhat similar, it may make sense to
	place them in similar orientations so that the average of all objects bears
	some resemblance to some low resolution version of the objects. 
	The first column of $\tilde{W}$ can then be thought of as representing the
	average object, and subsequent columns can be thought of as a coefficients 
	representing the variations from the average (e.g., SVD/PCA of the hyper-object).

        Next, we consider the following accounting exercise.
        Suppose that we $W$ is of dimensionality $M \times K$ ($M$ coefficients in the representation of
        each of the $K$ objects).
        Suppose that instead of arranging the coefficients in a matrix, we concatenate them
        into the vector $\mathbf{w}$:
        \begin{equation}
          w_{kM + i} = w^{(k)}_i,
        \end{equation}
        with $k=0,1,...,K-1$ and $i=0,1,...M-1$.
        Then
	\begin{equation}
          \begin{split}
		\hat{\mathcal{M}}((\omega_1,\omega_2,\omega_3),k) =& \sum_i w_{i}
		\varphi_{(i \text{~mod~} M)}(\omega_1,\omega_2,\omega_3) \delta\left(\lfloor {i/M}\rfloor,k\right) \\
                  =& \sum_i w_{i} \psi_{i}(\omega_1,\omega_2,\omega_3,k) ,
          \end{split}
	\end{equation}
        where
        \begin{equation}
          \psi_{i}(\omega_1,\omega_2,\omega_3,k)  = \varphi_{(i \text{~mod~} M)}(\omega_1,\omega_2,\omega_3) \delta\left(\lfloor {i/M}\rfloor,k\right) .
          \end{equation}
        In other words, by introducing the basis functions $\phi_i$, we represent the multiple objects as hyper-object,
        with an expansion that is analogous to Equation (\ref{eq:rep:het:disc}).

	\subsection{The representation of heterogeneous objects - the continuous case}

        Many existing software packages represent objects as a discrete set of samples
        points in a 3-D grid, the natural generalization of this representation
        is a higher dimensional grid, where some axes capture the heterogeneity.
        This is very similar to treating continuously heterogeneous molecules
        as if they were a collection of independent discrete states.
        The idea of this manuscript is to take advantage of relation between these
        states, some (but not all) of the approaches that we propose to implement this idea
        use implicit regularization through basis functions; the purpose of this section is
        to discuss the representation of hyper-objects.

       	Suppose that we have a family of objects, parameterized by the variable $t$, and
	continuous (in the appropriate way) with respect to $t$.
	In other words, for every value of $t$, we have an objects instance $\mathcal{V}[t]$ of the hyper-object, 
	and a small difference $d(t,t')$ between parameters implies a small
        difference between the instance objects, i.e.
	a small $d\left(  \mathcal{V}[t] , \mathcal{V}[t'] \right)$ (in an
	appropriate definition of distance).

        The natural generalization of (\ref{eq:rep:expansion}) to this
        higher dimensional case is:
	\begin{equation}\label{eq:multivolumeexpansion}
		\mathcal{M}((x_1,x_2,x_3),t) = \sum_i w_{i} \psi_i(x_1,x_2,x_3,t) ,
	\end{equation}
        with basis functions $\psi_i$ over $\mathcal{X} \times \mathcal{T}$,
        or, equivalently, in the Fourier domain
	\begin{equation}\label{eq:multivolumefourierexpansion}
		\hat{\mathcal{M}}((\omega_1,\omega_2,\omega_3),t) = \sum_i w_{i} \hat{\psi}_i(\omega_1,\omega_2,\omega_3,t) .
	\end{equation}
        One convenient way to produce such basis function
        is a tensor product between some standard 3-D object basis functions
        and basis functions appropriate for the parameter space,
        \begin{equation}\label{eq:rep:tensorfunctions}
          \psi_{ij} = \varphi_i(x_1,x_2,x_3) \eta_j(t),
        \end{equation}
        so that 
	\begin{equation}\label{eq:multivolumetensorexpansion}
	  \mathcal{M}((x_1,x_2,x_3),t) =  \sum_j \eta_j(t) \sum_i  w_{ij} \varphi_i(x_1,x_2,x_3) 
           =  \sum_i  \varphi_i(x_1,x_2,x_3) \sum_i  w_{ij} \eta_j(t) .
	\end{equation}	
        Appropriate choices of bases provide implicit regularization for reconstruction algorithm,
        so the basis functions $\psi_i$ should be chosen considering the types of objects and
        heterogeneity that we expect to model.

        In fact, one such generalization of objects to hyper-objects is a natural
        generalization of the standard use of the sample points in the Fourier domain.
        A 3-D object $\mathcal{V}$ is typically represented in cryo-EM algorithm using samples on grid points
        of the 3-D Fourier transform of the object $\mathcal{F}_X \mathcal{V}$.
        One of the ways to formulate a simple instance of hyper-objects, is to
        consider the 4-D hyper-object $\mathcal{M}$ and to represent is using
        sample points on a 4-D grid of the 4-D Fourier transform of the hyper-object
        $\mathcal{F}_X \mathcal{F}_T \mathcal{V}$,
        where the 4-D Fourier transform is the composition of the Fourier transform on the
        spacial axes $\mathcal{F}_X$ and the Fourier transform in the heterogeneity/``time''
        axis $\mathcal{F}_T$ (the number of sample points on each axis can be different from the number of sample points on other axes).
        Slightly more generally, $\mathcal{F}_T$ can be replaced with some arbitrary other
        transform $\mathcal{L}_T$ (see Section \ref{sec:analysis:het:hyper}).

	\subsection{Unknown orientations and heterogeneity parameters, and reconstruction
	of hyper-molecules in cryo-EM}
	
        The advantage of this general form of expressing hyper-objects
        is that the general linear operator formulation (see (\ref{eq:operatortomography}))
        applies with no conceptual change. An image is produced by the operation
        \begin{equation} 
	  \mathbf{y^{(i)}} = A(R_i, t_i) \mathbf{w} ,
	\end{equation}
        where $\mathbf{w}$ is a vector of coefficients that represents a hyper-object
        (as sample points or in some other basis) and $A^{(i)} = A(R_i, t_i)$ is an operator
        that produces an image at the correct orientation $R_i$ and parameter value $t_i$.
        While this is very similar to the formulation of discrete heterogeneity (see (\ref{eq:het:operatortomography})),
        this formulation highlights the idea of treating the heterogeneity parameter
        in the same conceptual way as the rotation parameter (to the extent possible).

        The traditional direct reconstruction algorithm is again a generalization of
        (\ref{eq:refine:orientation}) and (\ref{eq:refine:volume}):
        \begin{equation}\label{eq:hyperrefine:orientation}
          (R_{(n,i)},t_{(n,i)}) = \argmin_{(R,t)} \|A(R,t) \mathcal{M}^{(n)} - I^{(i)}\| ,
        \end{equation}
        and
        \begin{equation}\label{eq:hyperrefine:volume}
          \mathcal{M}^{(n+1)} = \argmin_{\mathcal{M}} \sum_{i = 1}^N \|A(R_{(n,i)},t_{(n,i)}) \mathcal{M} - I^{(i)}\|^2 .
        \end{equation}   
        What makes it different from  (\ref{eq:refine:het:orientation}) and (\ref{eq:refine:het:volume})
        is that $t$ can be continuous (or finely discretized), and that it reconstructs the
        hyper-object $\mathcal{M}$ in a way that uses relations between the object instances at
        different values of $t$.
        For example, it is possible to have a different value of $t_i$ for each image (in fact, this is
        the realistic model since $t_i$ is a continuous parameter, the differences between close points would
        of course have to be small); traditionally, the values of $t$ would either be restricted to a discrete
        number of unordered values that correspond to existing classes, or this would mean that  each image
        would be assigned to a separate class, making it impossible to reconstruct that class.
        At the same time,
        nothing would have been known about intervals where no samples had been taken.
        Indeed, in this generality, the reconstruction of a hyper-object appears to be
        severely ill-posed, but so is the reconstruction of a single  object.
        Loosely speaking, the same conceptual problem exists in the homogeneous case
        (with somewhat different properties): each image  has its own rotation
        so that most points in the Fourier space are never sampled more than once or twice.
        The reconstruction of non-heterogeneous objects is achieved using additional implicit and explicit properties, priors,
        regularization and penalties. 
        We propose to generalize the existing algorithms and to apply similar principals to the generalized problem.
        In this section we briefly review some of the approaches for generalization
        that we find of particular interest for continuous heterogeneity.
        The same general concepts apply to more elaborate approaches and algorithms used for
        reconstruction of objects in the homogeneous case and in the case of discrete heterogeneity.

        The following are some approaches to explicit or implicit regularization and priors.
        These approaches can be applied to discretized sampled states
        (connecting the currently independent states) or to more continuous
        representations of states (indeed, to some extent, these are equivalent, see Section \ref{sec:setup:reps}).
        The thread that connects these methods is the relation between states, or ``continuity.''
        Some of these approaches
        are standard constructions once the problem is stated as proposed in this paper,
        and some are less standard.
        The idea behind all these methods is to capture the relation between different
        ``states'' (heterogeneity parameter values) of the hyper-object instead of
        treating them as independent entities.
        \begin{itemize}
        \item Representing hyper-objects using tensor products of standard 3-D object basis functions and parameterization function (space or frequency domains):
          \begin{equation}
            \psi_{i,l}(\omega_1,\omega_2,\omega_3, t) = \varphi_i(\omega_1,\omega_2,\omega_3, t) \cdot \eta_l(t) 
          \end{equation}
          for all combinations of $i,l$.
        \item A similar tensor product, with restrictions on the combinations of $i,l$ allowed. For example, allowing high frequency object functions (indexed by $i$) to be combined with more (or fewer) parameterization functions (indexed by $l$).
        \item Penalties on some components, such as coefficients of high frequency heterogeneity functions.
        \item High dimensional functions of both space/frequency and parameter, and in particular, spaces of basis functions that are chosen to better capture predicted types of structure and variability (e.g., superimposing a homogeneous global representation of an object with a localized basis that allows variability in spatially restricted certain areas of the object).
        \item Using frames (e.g., high dimensional wavelet frames),  which capture the structure and variability (possibly with sparsity constraints).
        \item Optimal Transportation distance, Earth mover's distance (EMD) or Wasserstein distance (see discussion below).
        \item Sparsity in the use of basis function or in the spacial objects.
        \item Total variation (TV) regularization in the spacial domain or a generalized form.
        \item Priors or constraints on the distribution of images in the heterogeneity parameter space (and combinations of heterogeneity parameter and orientations).
        \item Tree or multiscale heterogeneity structures  (see below).
        \item Hyper-frequency marching (see Section \ref{sec:alg:freqmarch} below).
        \item Continuity through image assignment rather than object representation: soft assignment of class and rotation, which is aware of class topology, so that an image that is assigned probability to be certain class and rotation is also assigned probability to be in nearby classes with similar rotations (up to global rotations).          
        \item Adapting continuity constraints and regularization methods developed for 4-D CT (e.g., \cite{jia20104d,gao20124d,kazantsev2015employing,gopi2014multiple,low2005novel5d}) with appropriate variations and generalization to more arbitrary parameter spaces.
          \end{itemize}

	We note that a source of true continuous heterogeneity in cryo-EM is
	flexibility of the molecule. To the extent that the molecule can be modeled as a
	sum of smaller objects (atom or substructures), it may be useful to regularize
	the variability in the hyper-object using a penalty based on EMD 
	$d_{EM}$ (see, for example, \cite{rubner2000earth}), so that for small $d(t,t') < \delta$, the EMD
	$d\left( \mathcal{V}[t] , \mathcal{V}[t'] \right)$ between the objects instance
	$\mathcal{V}[t]$ at parameter value $t$ and the nearby object distance
	$\mathcal{V}[t']$ at parameter value $t'$ is small.
	Loosely speaking, EMD distinguishes between local
	variability in distributions of masses and more global redistributions of
	mass, so it may be suited to capture local variability in the location of
	sub units.
        Extensions of this treatment of distances provides, for example, methods for
        interpolating between instances.

	The tree construction could potentially have several interesting properties.
        In its simplest form, the tree construction uses parameter basis functions
        that decompose the parameter space into ``intervals,'' in each interval
        the function is constant.
        Technically speaking, and in the absence of additional constraints,
        this construction appears to be no different than a standard decomposition
        to multiple discrete classes practiced in existing algorithms.
        Furthermore, it is essentially equivalent to using the Haar basis (see (\ref{eq:haar}))
        for the parameter expansion, which yields non-continuous functions,
        but otherwise falls nicely into the proposed approach to expand in arbitrary bases.
        However, when regularization, such as high-frequency component penalties, is introduced,
        the different instances are tied together.
        Indeed, this construction does not offer a clear continuous parameterization
        of the heterogeneity, but, it could be less sensitive to the choice
        of parameterization topology and expansion (at the cost of not being
        adapted to using prior knowledge of the appropriate topology and
        expansion).
        This approach can be extended with additional multiscale representations;
        preliminary results indicate that adding non-constant basis functions
        to the expansion are likely to contribute to the construction.

        Due to the limited scope of this first discussion of this approach,
        our discussion of the first prototype algorithm is focused on implicit regularization
        through the choice of basis functions and frequency marching. 
	For example, if the parameter space is $\mathcal{T} = [0,1]$,  
	we use in the tensor product (see (\ref{eq:multivolumetensorexpansion}))
        of standard bases used for objects, and 
        and a family of polynomials in the heterogeneity axis (for more details, see
	Section \ref{sec:alg}).

	\subsection{A remark on ambiguity}\label{sec:amb}

        As is well-known, there are multiple equivalent solutions to classic cryo-EM
        reconstruction problems due to certain symmetries.
        For example, classic cryo-EM cannot distinguish between a molecule and its mirror
        image, so two mirror solutions are considered equivalent.
        Similarly, any solution is unique only up to rotation of the molecule.
        Furthermore, in the case of discrete heterogeneity, the solution is unique only
        up to permutations of the classes (i.e. any of the reconstructed molecule can be
        called ``molecule number 1''), and rotation of each class independently of the others.

        The reconstruction of hyper-objects is subject to some similar ambiguities.
        For example, a hyper-object is generally equivalent to a rotated version of itself or a reflected
        version.
        In addition, a hyper-object $\tilde{\mathcal{M}}$ created by rotating the object $\mathcal{M}$ to a different
        orientation $R(t)$ at every value of the heterogeneity parameter $t$, as define by the
        formula
        \begin{equation}
          \tilde{\mathcal{M}}[t] =  R(t) \mathcal{M}[t] ,
          \end{equation}
        would be equivalent to  $\mathcal{M}$. Typically, $R(t)$ would be continuous with
        respect to the parameter space $\mathcal{T}$ if we assume a continuous representations.
        Furthermore, in the absence of a unique metric on the parameter space  $\mathcal{T}$,
        the hyper-object is subject to reparameterization. For example, suppose that the parameter space is
        one dimensional: $\mathcal{T} = [0,1]$, then the reparameterized hyper-object $\tilde{\mathcal{M}}$
        defined by the formula
        \begin{equation}
          \tilde{\mathcal{M}}[t] =  \mathcal{M}\left[ f(t) \right] ,
          \end{equation}
        where $f$ is an increasing function on the interval with $f(0)=0, f(1)=1$,
        is equivalent to the original hyper-object. In some cases, it may be possible to
        regularize the hyper-object or introduce a metric that reduces the ambiguity
        in the parameterization, for example by requiring a parameterization
        where the samples are uniformly distributed in the parameter space,
        or penalizing high order heterogeneity coefficients.

        We note that some ambiguities can technically yield hyper-objects that
        are not equivalent. For example, it is technically possible for an
        algorithm to use the degrees of freedom provided by the hyper-object
        model to capture shifts in images instead of an interesting variation
        (much like it is technically possible that some existing algorithm
        would generate two molecules that are rather similar shifted copies of one another). 
        Such ambiguities and parameterization considerations are likely to be
        of interest in more elaborate implementation of the approach proposed in this manuscript.

%
%
%
\section{Algorithms}\label{sec:alg}

The approach presented in this manuscript is quite general, and can be
used to generalize various algorithms.
The purpose of this section is to briefly present a simplified algorithm
which we have implemented as proof-of-concept for our approach.
Since this particular algorithm is only one of many ways to implement our
approach, we will present it in limited detail.
We defer the more comprehensive discussion of the ideas introduced in this section
to future papers.

	\subsection{Expansion of hyper-objects}

        We represent an object using concentric shells in the Fourier domain, 
        with the function on each shell expanded in spherical harmonics,
        as in \cite{barnett2016rapid}.
        The $k$-th shell of an object $\mathcal{V}$ is represented using the expansion
        \begin{equation}\label{eq:alg:vol_shell_rep}
          \mathcal{V}( k , \theta, \phi ) =  \sum_{n=0}^{p(k)} \sum_{m=-n}^n v_{k,n,m} Y_n^m (\theta,\phi) ,
        \end{equation}
        where $\theta$ and $\phi$ define a position on the sphere, $Y_n^m$ are spherical harmonics (see (\ref{eq:sphericalharmonics})),
        and $v_{k,m,n}$ are coefficients of the expansion.
        For each image we compute the Fourier expansion of concentric
        circles in the Fourier domain. The $k$-th circle in the Fourier transform of the $i$
        image is expanded as follows:
        \begin{equation}\label{eq:alg:image_circ_rep}
          \hat{I}^{(i)}( k , \phi ) =   \sum_{m=-p(k)}^{p(k)} \alpha^{(i)}_{k,m} exp(i m \phi) .
        \end{equation}
        Unlike the implementation in \cite{barnett2016rapid},
        all our operations are computed directly with spherical harmonics
        using rotation operators that produce the coefficients of a rotated sphere
        from the coefficients of a sphere.
        The rotations of the sphere and the restriction of the sphere
        to the circles are sparse operations
        (see (\ref{eq:spherecoeffrot}) and (\ref{eq:sphererestrict}), respectively)
        in the sense that they mix very few of the coefficients in the expansion
        of the object.

        Our preliminary implementation of hyper-objects allows one dimensional
        parameterization over the interval $[0,1]$.
        We expand the hyper-object $\mathcal{M}$ in a basis that is the tensor product
        between the basis for objects in (\ref{eq:alg:vol_shell_rep}) and
        heterogeneity basis functions which we denote by $\overline{P_q}$;
        here we chose either normalized Legendre polynomials or Chebyshev polynomials (see, inter alia, \cite{abramowitz1964handbook}),
        shifted to the appropriate interval:
        \begin{equation}\label{eq:alg:multivol_shell_rep}
          \mathcal{V}( k , \theta, \phi, t ) =  \sum_{q=0}^{Q} \sum_{n=0}^{p(k)}
                 \sum_{m=-n}^n v_{q,k,n,m} \overline{P_q}(t) Y_n^m (\theta,\phi) .
        \end{equation}

	\subsection{``Hyper-frequency marching''}\label{sec:alg:freqmarch}

        The algorithm implemented here generalizes the idea of frequency marching
        presented in  \cite{barnett2016rapid},
        where the expansion is initially restricted a small number of shells,
        with later iterations allowing a growing number of shells in the expansion
        (larger values of $k$ allowed in later iterations).
        Our implementation follows a similar approach, increasing the allowed $k$ in later iterations,
        however, we also restrict the parameterization basis; we start with $Q=0$ (no heterogeneity),
        and later increase the number $Q$ of basis functions allowed in the expansion.

	\subsection{A simplified stochastic optimization algorithm}

        The algorithm implemented here is based on Stochastic Gradient Descent (SGD).
        An SGD algorithm for the Bayesian approach to cryo-EM has been proposed in \cite{punjani2017cryosparc},
        our simplified implementation is an SGD version of a more traditional algorithm,
        with the continuous heterogeneity generalization proposed in this paper. 
        At each iteration of the algorithm, we choose a small number of images (a ``mini-batch''),
        and estimate the orientation and heterogeneity parameter of each image.
        We then compute the gradient of the object that would decrease the discrepancy between
        the images in the mini-batch and the object at the selected orientation,
        and make a small update to the object in that direction.

        \subsection{Sampling}

        Our algorithm requires samples the continuum of possible rotations and heterogeneity
        parameter values in order to generate templates at every iteration.
        In the current implementation, we sample both the rotations and parameter values randomly,
        with the exception of in plane rotations of the template,
        in which the computation of the cost function is accelerated
        using FFT as discussed in  \cite{low2003method,barnett2016rapid}.
        
        The acceleration technique in \cite{low2003method,barnett2016rapid}
        generalizes to allow faster computation
        of the cost function for different values of
        the parameter $t$ (not yet implemented in this example implementation, however tested in other versions).

%
%
%

\section{Experimental Results} \label{sec:res}

The simplified algorithm discussed above was implemented in Matlab.
This simplified proof of concept does not consider shifts in the images,
CTF, etc., but it does included simulated noise.

\begin{figure}[t!]
\begin{center}
  \includegraphics[width=0.45 \linewidth]{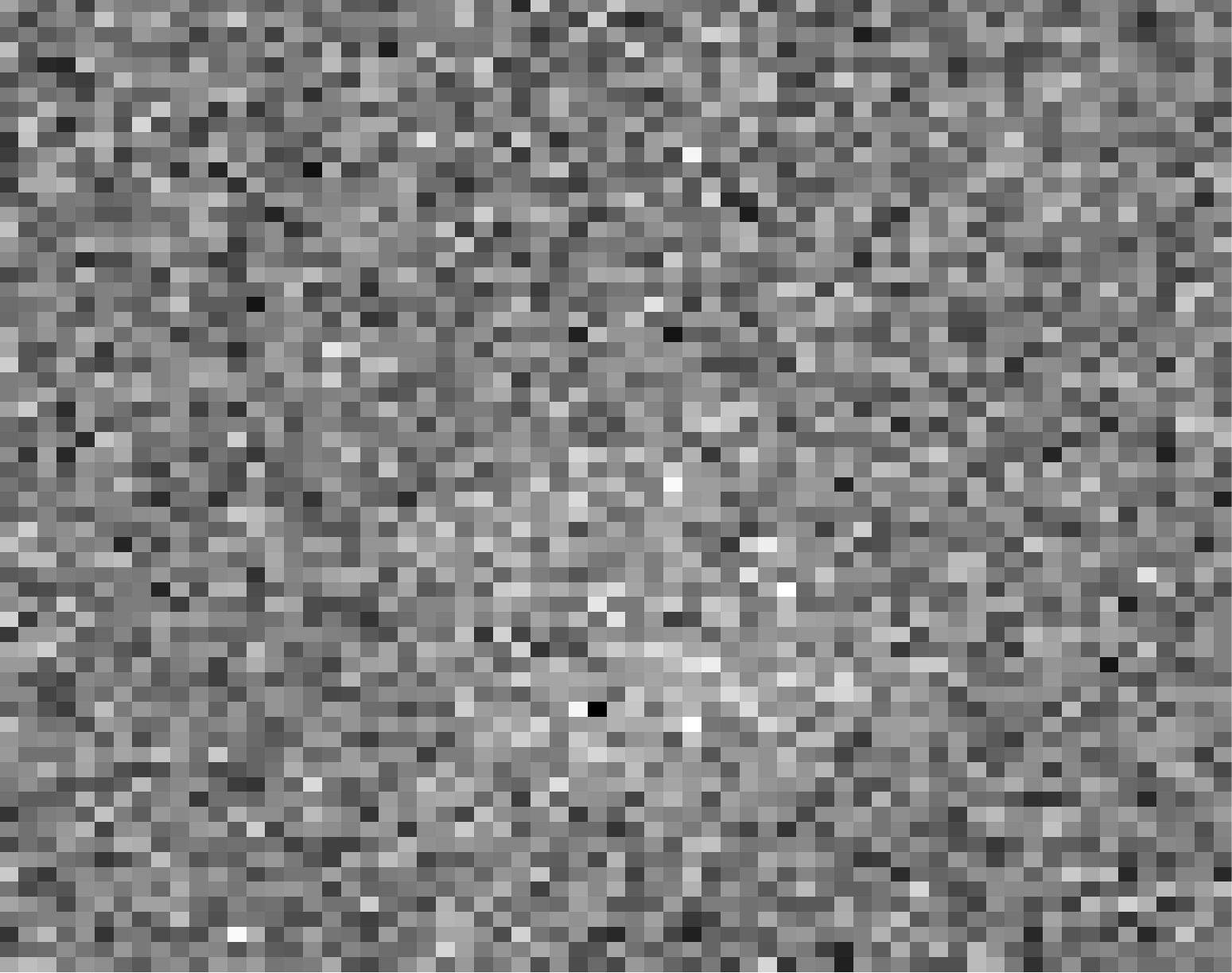}%
  ~
  \includegraphics[width=0.45 \linewidth]{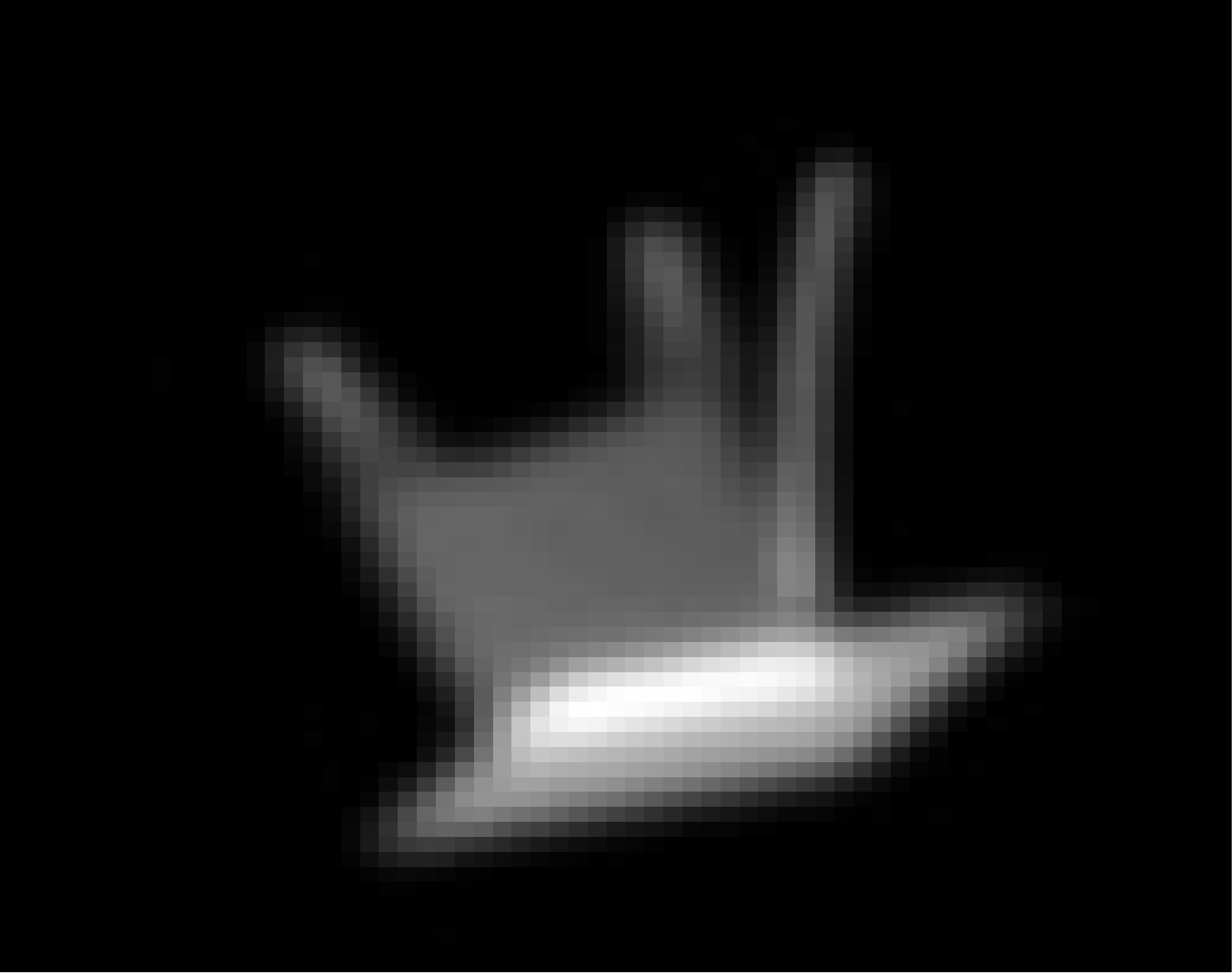}%
\end{center}
\caption{Sample of a noiseless projection image (right) and the same image with noise (left), as used in the simulation } \label{fig:sim:images}
\end{figure}

We simulated a hyper-object (a ``cat'') composed of Gaussian elements
in real space, where each Gaussian follows a continuous
trajectory as a function of the parameter, so that 
we have a continuous space of objects.
Examples of simulated 3-D objects instances are presented in
Figure \ref{fig:sim}.
The hyper-object displays large scale extensive heterogeneity
in the form of flexibility which is difficult to model as
a combination of small number of rigid objects.
We used the simplified imaging model
(see (\ref{eq:cryoemmodel:space})) to simulate $65\times 65$ pixels
images in various orientations and parameter values,
and added Gaussian noise to the images.
The SNR in this experiment was $1/16$.
Examples of the simulated images are presented in Figure \ref{fig:sim:images}.

In our preprocessing stage, we compute the
Fourier transform of the image at sample points on concentric circles in the
Fourier domain, and then use FFT to obtain the image coefficients on concentric circles
defined in (\ref{eq:alg:image_circ_rep}).

We run the algorithm ab initio, without an initial guess. 
We initialize the hyper-object to a non-heterogeneous spherically symmetric object,
by setting each sphere to the average of the corresponding circles
across all images, i.e. we set the $0$-th order coefficient in the
expansion of the $k$-th shell to the normalized average of 
the $0$-th order coefficients $\alpha^{(i)}_{k,0}$ of the $k$-th circle
of all the images.

We use SGD iterations to update  the non-heterogeneous model object
and after several steps of frequency marching 
we alternate between increasing the frequency limit $K$ and the number of
parameterization basis functions $Q$.

For the purpose of visualization of the computed hyper-object,
we choose sample values of the parameter $t$,
and then sampled the shells at quadrature points in the Fourier domain
and used NUFFT (see \cite{greengard2004accelerating})
to recover each object  at regular grid points in real space.
We use Matlab's ``isosurface'' and ``patch'' to visualize level sets of the objects. 

The algorithm was run on computer equipped with Intel(R) Core(TM) i7-4770 CPU,
32GB RAM server and a single NVIDIA GeForce GTX 980 Ti GPU with 6 GB GPU RAM
for about 5 hours. We note that very similar results have been obtained
even in the 1-2 hours of this experiment and of similar shorter experiments. 

Examples of reconstructed objects are presented in Figure \ref{fig:res}.
We observe that the reconstructed objects appear to be slightly rotated
one with respect to the other compared to the simulated data,
due to the ambiguity discussed in Section \ref{sec:amb}.
In fact, the relative rotations in the result may reflect a better
choice than our choice in the simulation, in terms of the rate of change
in the hyper-object, or in terms of the norm of the heterogeneity coefficients.

\begin{figure}[h!]
\begin{center}
\includegraphics[width=0.75 \linewidth]{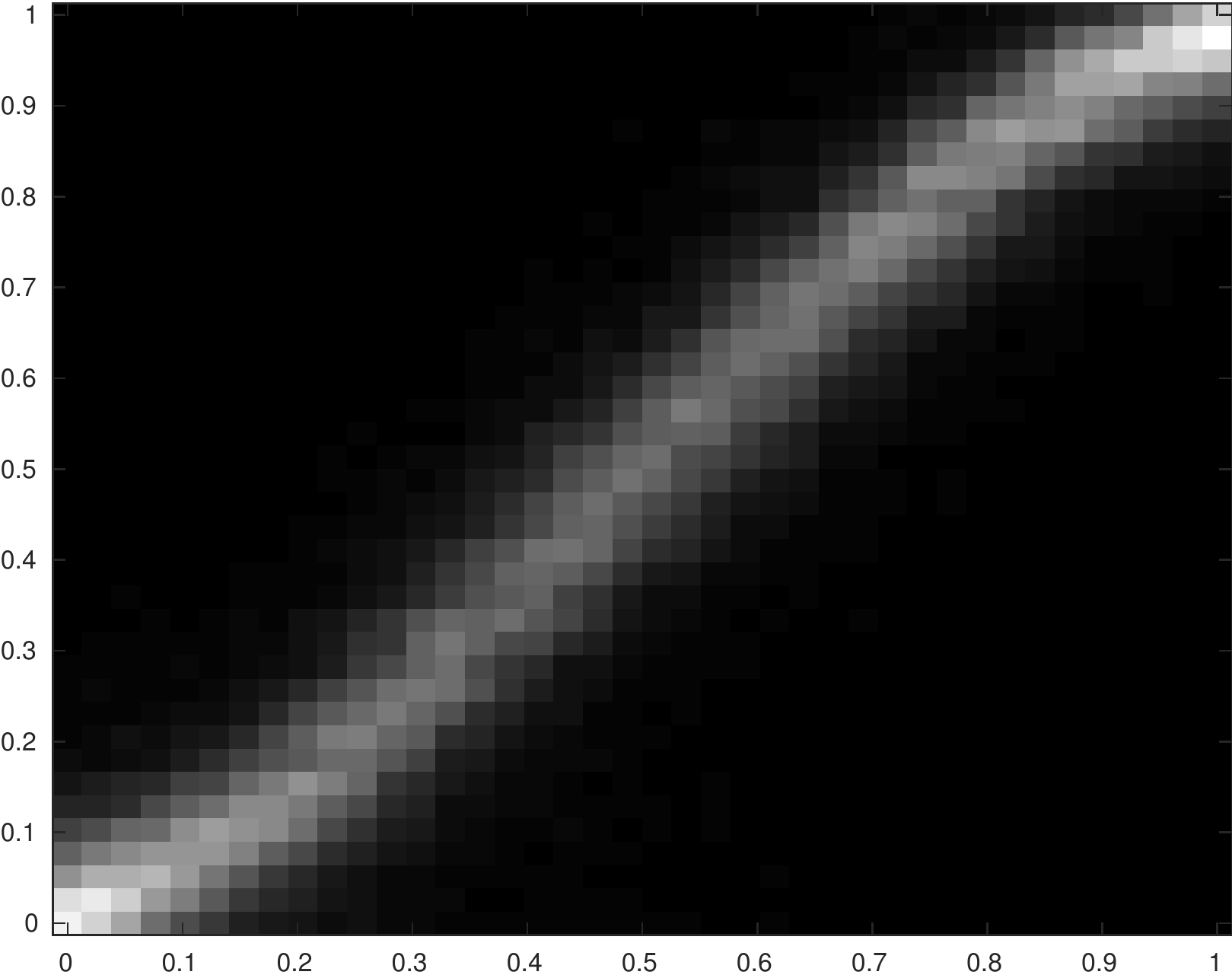}%
\end{center}
\caption{Distribution of the pairs true parameter values of an image (x-axis), and estimated parameter values (y-axis) in the parameterization that the algorithm finds, as assigned by the algorithm during the final iterations} \label{fig:mapmain}
\end{figure}

\begin{figure}[h!]
\begin{center}
\includegraphics[width=0.75 \linewidth]{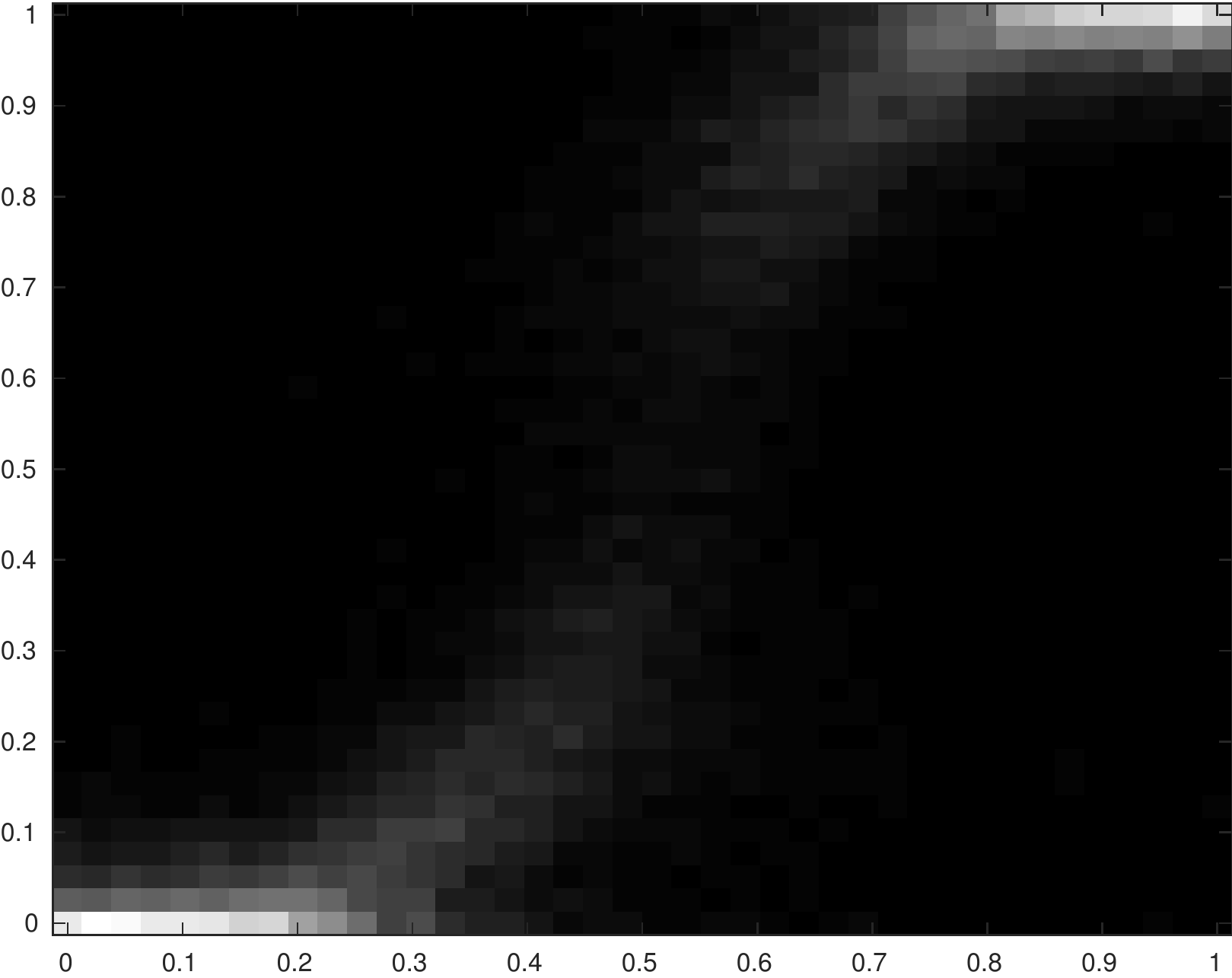}%
\end{center}
\caption{Distribution of the pairs true parameter values of an image, and estimated parameter values in the parameterization that the algorithm finds in an experiment with fewer iterations and more rapid marching. Here, the distribution is not uniform and only part of the heterogeneity is captured.} \label{fig:map:disc}
\end{figure}

\begin{figure}[h!]
\begin{center}
\includegraphics[width=0.75 \linewidth]{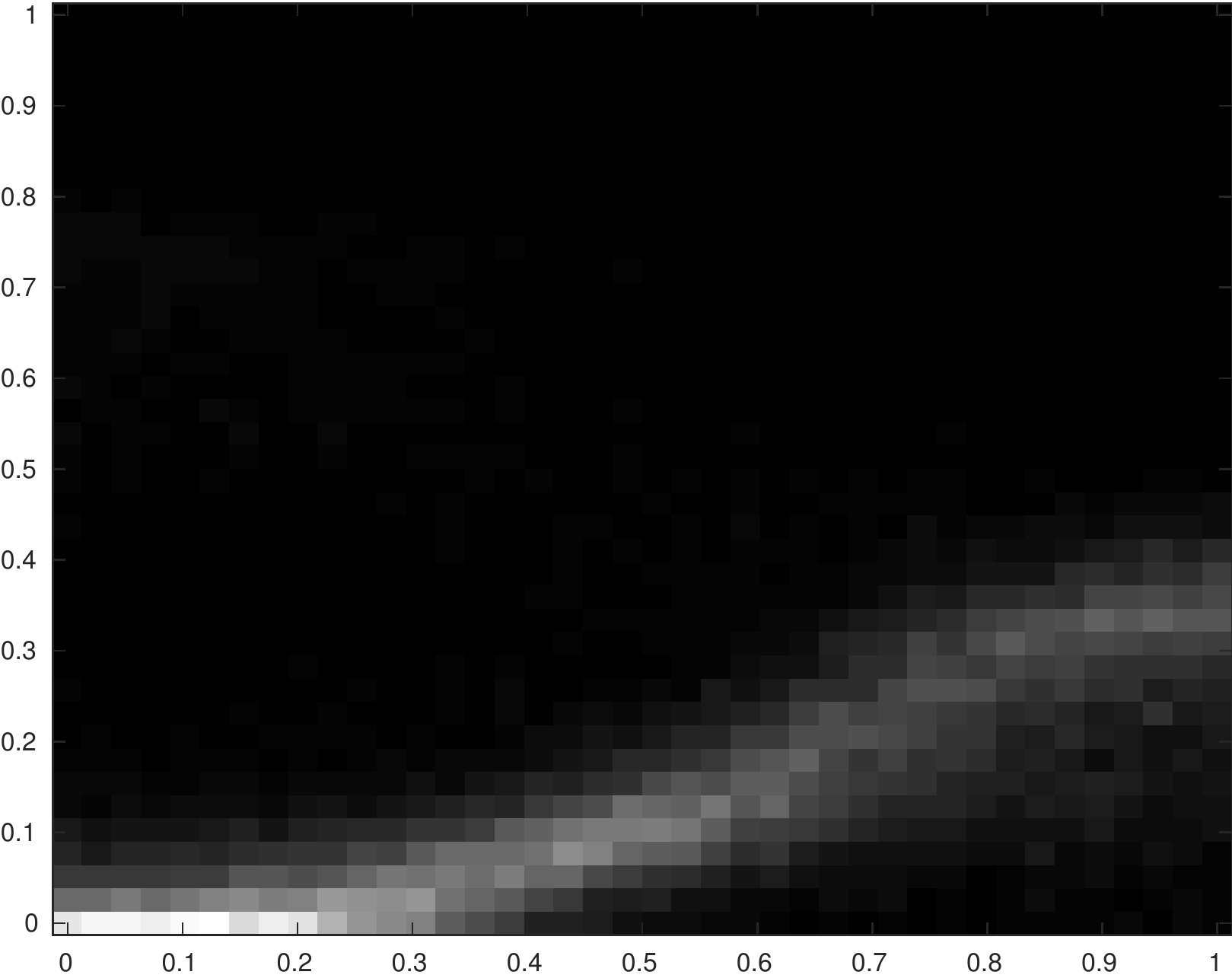}%
\end{center}
\caption{Distribution of the pairs true parameter values of an image, and estimated parameter values in the parameterization that the algorithm finds in another experiment with fewer iterations and more rapid marching. Here, only part of the parameter space is used.} \label{fig:map:part}
\end{figure}

In Figure \ref{fig:mapmain}, we present the distribution of the pairs of
true-parameter (x-axis) and estimated parameter (y-axis) assigned to images during  the
final steps of the refinement, before the algorithm was stopped. 
This current simple implementation does not regularize or reparameterize the parameter space,
and has been used ab initio, without any initial estimate;
while it succeeds in obtaining an appropriate parameterization in the surprising majority of the
runs, in some runs, typically when using a small number of iterations or very rapid frequency marching in early stages of the run,
we find some examples of inefficient parameterization, which is typically not difficult to detect.
For example, in Figure  \ref{fig:map:disc} we find that the algorithm maps most of the parameter
space to extreme points, preliminary results suggest that even this naive implementation of the algorithm
gradually reparameterizes this mapping, but this process is relatively slow.
This incident is relatively easy to detect by examining the marginal distribution on the y-axis
(under the assumption or constraint of a more uniform distribution). 
Another example, shown in Figure \ref{fig:map:part} demonstrates a way in which the algorithm
can use only part of the allowed parameter space. The reconstruction results are still good in the region of
space that the algorithms chose to use, but it is an inefficient use of the parameter space.
This case is also easy to detect and correct.
Conservative optimization appear to mitigate these issues even in the current implementation; 
initial estimates of the objects and better regularization are also likely to mitigate these issues
when they occur.

Video visualizations of the results are available at the website \\
\url{http://roy.lederman.name/cryo-em}.

~\\

\clearpage	
 
%
%
%
\section{Summary and future work}\label{sec:future}

The idea behind this work is to extend the existing tools in cryo-EM
to the case of continuous heterogeneity.
Conceptually, we attempt to treat the heterogeneity parameter (or ``class'')
in the same way as the orientation.
We propose various approaches for representing the heterogeneous hyper-objects
and for regularizing the problem.
In this manuscript, we presented one simple expansion of the hyper-object and one
simple algorithm and implementation of these ideas as proof-of-concept.
In this section we discuss some future directions for this line of work.

We believe that various versions of this general approach have several advantages compared to
the construction of independent classes:
it allows to use assumptions on the topology and properties of the heterogeneity in a given problem
to improve the reconstruction,
it captures the topology of the heterogeneous structures and provides a natural way to
reconstruct object instances at arbitrary parameter points,
it combines information from different nearby object instances to produce
a fine-grained spectrum of states, leveraging continuity assumptions.

While many of the representation and regularization techniques
proposed here can be implemented using sampled discretized
values of the heterogeneity parameter, we find is particularly useful
to use continuous bases in this discussion, because they offer 
a natural generalization of existing algorithms. 
In this manuscript we implemented only several
continuous expansion using a tensor product of bases of functions
and standard bases for the expansion of 3-D objects.
More elaborate bases and other continuous and discrete
representations of hyper-objects are likely to be useful
in capturing properties of the heterogeneity and further development of this approach.

One possible source for methods that could be adapted, is the existing body
of work in 4DCT. 
Since 4DCT typically treats the case where both the orientation and heterogeneity parameter value
(phase in the breathing cycle) are known, not all the ideas in 4DCT are directly applicable.
At the same time, the investigation of unknown parameters in cryo-EM may contribute to methods in CT,
in cases where the breathing phase is not recorded accurately, or when there 
are additional heterogeneity parameters.

The examples used in this manuscript included only one dimensional
heterogeneity on an interval, however the approach is general
and can be used to investigate more elaborate topologies of heterogeneity.
The investigation of other topologies and the sensitivity to the choice of
topology, metrics, and bases is another direction in this line of work. 
We note that since the algorithms are relatively fast, it is possible to
try multiple topologies, and we note that the tree/Haar approach is likely to
be less sensitive to the precise choice of topology.
However, the choice of topology is a useful implicit regularizer or prior.
Another aspect of the work on topology is to develop methods of detecting local problems
in the parameterizations that algorithms discover.
While our discussion has been devoted mostly to continuous heterogeneity,
the methods are applicable to discrete heterogeneity, for example
when there are small changes between two conformations.
A related matter that merits additional investigation is the case where
some states in the continuum of states are far more represented than others,
and the related issue of defining a metric on the space of heterogeneity.

The approach is general and we submit that it can be used in various existing cryo-EM algorithms
(although the actual software implementations may require significant modifications).
It is possible to combine existing software with this approach in several ways,
for example, by crude continuous classification of images in this approach, followed by a local reconstruction,
or a reconstruction of an initial homogeneous model of the molecule,
and then superimposing local basis functions in areas of variability
in the model and running this type of algorithm to resolve the heterogeneity.

While the results obtained using this preliminary prototype are promising, 
additional work is needed to complete the implementation of this approach
in an algorithm that is efficient enough to
reconstruct high resolution hyper-objects in real cryo-EM settings,
while utilizing modest computational resources.

%
%
%
\section{Conclusions}\label{sec:conclusions}

A framework has been presented for the tomography inverse problem
in the case of continuously heterogeneous objects,
where the orientation and heterogeneity parameter values are unknown.
The proposed framework generalizes existing approaches for the
reconstruction of molecules in cryo-EM,
so that, in principal, existing algorithm can 
be adapted to the case of continuous heterogeneity.

The approach has been demonstrated in simplified simulated
cryo-EM settings, using one of the proposed implementations,
and a new prototype algorithm.

We are currently working on expanding our investigation of
alternative representations and regularization approaches
within this framework, and adapting the implementation to
real-world cryo-EM applications.

\section*{Acknowledgments}

The authors were partially supported by Award Number R01GM090200 from the NIGMS, FA9550-12-1-0317 and FA9550-13-1-0076 from AFOSR, Simons Foundation Investigator Award,  Simons Collaboration on Algorithms and Geometry, and the Moore Foundation Data-Driven Discovery Investigator Award.


\bibliography{bib08sub}{}
\bibliographystyle{ieeetr}


%

\end{document}